\documentclass{article}

\usepackage[english]{babel}
\usepackage{microtype}
\usepackage{graphicx}
\usepackage{booktabs} 
\usepackage{multirow}
\usepackage{array}

\usepackage{hyperref}

\usepackage[accepted]{icml2025}

\usepackage{amsmath}
\usepackage{amssymb}
\usepackage{mathtools}
\usepackage{amsthm}

\usepackage[capitalize,noabbrev]{cleveref}

\theoremstyle{plain}

\theoremstyle{definition}

\theoremstyle{remark}

\usepackage[textsize=tiny]{todonotes}
\usepackage{subfig}
\newcolumntype{C}[1]{>{\centering\arraybackslash}p{#1}}

\icmltitlerunning{Counterfactual Influence as a Distributional Quantity}

\begin{document}

\twocolumn[
\icmltitle{Counterfactual Influence as a Distributional Quantity}

\icmlsetsymbol{equal}{*}

\begin{icmlauthorlist}
\icmlauthor{Matthieu Meeus}{imperial}
\icmlauthor{Igor Shilov}{imperial}
\icmlauthor{Georgios Kaissis}{deepmind}
\icmlauthor{Yves-Alexandre de Montjoye}{imperial}
\end{icmlauthorlist}

\icmlaffiliation{imperial}{Imperial College London}
\icmlaffiliation{deepmind}{Google DeepMind}

\icmlcorrespondingauthor{Yves-Alexandre de Montjoye}{deMontjoye@imperial.ac.uk}

\icmlkeywords{Memorization, language models, machine learning}

\vskip 0.3in
]



\printAffiliationsAndNotice{} 

\begin{abstract}
Machine learning models are known to memorize samples from their training data, raising concerns around privacy and generalization. Counterfactual self-influence is a popular metric to study memorization, quantifying how the model’s prediction for a sample changes depending on the sample's inclusion in the training dataset. However, recent work has shown memorization to be affected by factors beyond self-influence, with other training samples, in particular (near-)duplicates, having a large impact. We here study memorization treating counterfactual influence as a distributional quantity, taking into account how \emph{all} training samples influence how a sample is memorized. For a small language model, we compute the full influence distribution of training samples on each other and analyze its properties. We find that solely looking at self-influence can severely underestimate tangible risks associated with memorization: the presence of (near-)duplicates seriously reduces self-influence, while we find these samples to be (near-)extractable. We observe similar patterns for image classification, where simply looking at the influence distributions reveals the presence of near-duplicates in CIFAR-10. Our findings highlight that memorization stems from complex interactions across training data and is better captured by the full influence distribution than by self-influence alone.
\end{abstract}

\section{Introduction}

Studying how Large Language Models (LLMs) learn from, and memorize, their training data is crucial to understand how they retain factual knowledge~\citep{kandpal2023large,petroni2019language} and generalize~\citep{wang2024generalization}, but also to assess privacy risks~\citep{carlini2021extracting,nasr2023scalable}, concerns related to copyright~\citep{meeuscopyright,karamolegkou2023copyright}, and benchmark contamination~\citep{oren2023proving,mirzadehgsm}. To quantify how a model memorizes a sample, \citet{zhang2023counterfactual} introduce \emph{counterfactual memorization} (self-influence), computing how much a model's prediction for a sample changes when that sample is excluded from training. Recent studies~\citep{shilov2024mosaic,liu2025language}, however, suggest that LLMs memorize substantially across near-duplicates, widespread in LLM training data. This raises the question whether memorization of a piece of text can be viewed in isolation, i.e. by just looking at the self-influence.

\textbf{Contribution.} We investigate memorization by considering how the \emph{entire} training dataset impacts the model's predictions on the target sample, rather than relying solely on self-influence. We adopt the framework from~\citet{zhang2023counterfactual} to compute \emph{counterfactual influence}, which measures how any training sample $x_i$ impacts the prediction for a target $x_t$. Rather than focusing only on self-influence (where $x_i = x_t$), we compute the influence of all training samples on all targets and treat this as a distributional property.

We conduct our analysis on GPT-NEO 1.3B~\citep{gao2020pile} models finetuned on the Natural Questions dataset~\citep{kwiatkowski2019natural}, computing the full influence matrix to quantify the contribution of each training point to each target prediction. We consider two types of records: (i) random, \emph{unique records} from the data distribution, and (ii) random samples for which we include artificially crafted near-duplicates in the training data (\emph{records with near-duplicates}). We find that unique records exhibit strong self-influence, whereas records with near-duplicates show more diffuse influence patterns; on average, their self-influence is $3$ times lower and spread across similar records. Yet, these records are significantly more \emph{extractable}; on average $5$ times more, suggesting that self-influence alone underestimates this key risk associated with LLM memorization. We then find the ratio of the largest to the second largest influence, \emph{Top-1 Influence Margin}, to clearly distinguish unique records from those with near-duplicates more effectively than self-influence. Finally, we study influence distributions for classification models trained on CIFAR-10~\citep{krizhevsky2009learning}, finding that the distribution alone exposes near-duplicates. 

Our results show that how models memorize their training data, especially in the presence of near-duplicates, might not be adequately captured by self-influence alone. Instead, memorization is multi-faceted phenomenon that can be better understood through the full distribution.

\section{Background}
\citet{zhang2023counterfactual} define counterfactual influence as the expected change in a model’s loss on a target example $x_t$ when a specific training point $x_i$ is included versus excluded from the training data. This formulation represents an extension from label memorization in
classification models as introduced by \citet{feldman2020neural} to language models. Formally, let $\mathcal{D}$ denote the underlying data distribution, and let $D \sim \mathcal{D}$ be a dataset sampled i.i.d.\@ from $\mathcal{D}$, the influence of $x_i$ on target record $x_t$ is:
\begin{equation}
    \mathcal{I}(x_i \Rightarrow x_t) = \underset{A_j: x_i \notin D_j}{\mathbb{E}} \left[\mathcal{L}_{A_j}(x_t)\right] - 
     \underset{A_j: x_i \in D_j}{\mathbb{E}}\left[\mathcal{L}_{A_j}(x_t)\right]
\label{eq:influence}
\end{equation}
where $A_j$ denotes a model trained on $D_j$, and $\mathcal{L}_{A_j}(x_t)$ is its loss on the target example $x_t$. Note that we adopt a sign change from the original definition, so that a positive value indicates that $x_i$ contributes positively -- reduces the loss -- to the prediction of $x_t$. \citet{zhang2023counterfactual} primarily study the case when $x_i = x_t$, which they define as the counterfactual memorization or self-influence of $x_t$, i.e. $\mathcal{I}(x_t \Rightarrow x_t)$. While all influence values in this work correspond to counterfactual influence, we omit the qualifier and simply refer to them as (self-)influence throughout.

Here we examine memorization taking into account the entire influence distribution (i.e. $\mathcal{I}(x_i \Rightarrow x_t)$ for all $x_i$) rather than solely considering the value of self-influence. We do so for both unique records and, in particular, when the underlying data distribution contains \emph{near-duplicates}. We say that $x_t'$ is a \emph{near-duplicate} of $x_t$ if the distance between them, measured by a chosen function $d(\cdot, \cdot)$, satisfies $0 < d(x_t, x_t') < \epsilon$ for some threshold $\epsilon > 0$. Examining Equation~\ref{eq:influence}, when the data distribution $\mathcal{D}$ contains many near-duplicates $x_t'$ of $x_t$, these duplicates are likely to be included in the sets where $x_t \in D_j$ and where $x_t \notin D_j$. If these near-duplicates strongly influence model behavior on $x_t$, the marginal effect of including $x_t$ itself might get diminished. In this work, we investigate how this manifests itself on the entire influence distribution.

\section{Related work}

\textbf{Extractability.} A key risk arising from LLM memorization is the possibility of \emph{extracting} training sequences from a model’s outputs~\citep{carlini2021extracting}. This phenomenon has hence been used to study memorization.~\citet{nasr2023scalable} distinguish between two types: \emph{discoverable} memorization, when prompted on a training data prefix, the model's greedy output exactly matches the corresponding suffix (as in \citealt{carlini2022quantifying}); and \emph{extractable} memorization, where training data can be generated from any prompt.~\citet{ippolito2023preventing} argues that exact matching is too strict, and propose \emph{approximate memorization} based on similarity metrics (e.g. BLEU).~\citet{hayes2025measuring} further extend this to stochastic decoding, quantifying the chance of generating a target sequence across multiple non-greedy samples.

\textbf{Membership Inference Attacks (MIAs).} Another approach to study memorization measures the performance of MIAs~\citep{shokri2017membership}, inferring whether a sample was seen during training. MIAs offer an effective measure of privacy risk, as defending against them also protects against stronger attacks like data reconstruction~\citep{salem2023sok}. Attacks range from white-box methods~\citep{wangmope} to black-box probes using model logits~\citep{yeom2018privacy,carlini2019secret,mattern2023membership,shidetecting}. MIAs have been studied for LLM pretraining~\citep{shidetecting,meeus2024did}, albeit with limited success~\citep{duanmembership}, and in finetuning~\citep{mireshghallah2022empirical}. Prior work has found that longer sequences repeated more often are more at risk~\citep{kandpal2022deduplicating,meeuscopyright}. Importantly for this work,~\citet{shilov2024mosaic} show that the presence of near-duplicates makes records more susceptible against MIAs. 

\textbf{Influence functions.} Introduced in robust statistics~\citep{hampel1974influence}, influence functions have been studied in the context of machine learning.~\citet{koh2017understanding} was the first to use influence functions for \emph{data attribution}, i.e. which training samples $x_i$ make the model generate its prediction for $x_t$ where $x_t$ is typically a held-out test sample. Since computing the true counterfactual (e.g., retraining without $x_i$) quickly becomes prohibitively expensive, they proposed an efficient second-order approximation using Hessian-vector products. Later work developed lighter-weight alternatives, such as TracIn~\citep{pruthi2020estimating}, which accumulates gradient-similarity scores over checkpoints. Recent work extends these methods to LLMs, to study generalization~\citep{grosse2023studying} or to estimate the value of individual records~\citep{choe2024your}. Here, we apply influence to study memorization by considering target records $x_t$ from the training data, using the counterfactual definition~\citep{zhang2023counterfactual} in a small-scale setting.

\section{Experimental setup}
\label{sec:experimental_setup}
\begin{figure*}[!t]
\centering
\subfloat[Influence matrix $I$\label{fig:influence_matrix}]{
    \includegraphics[width=0.2\textwidth]{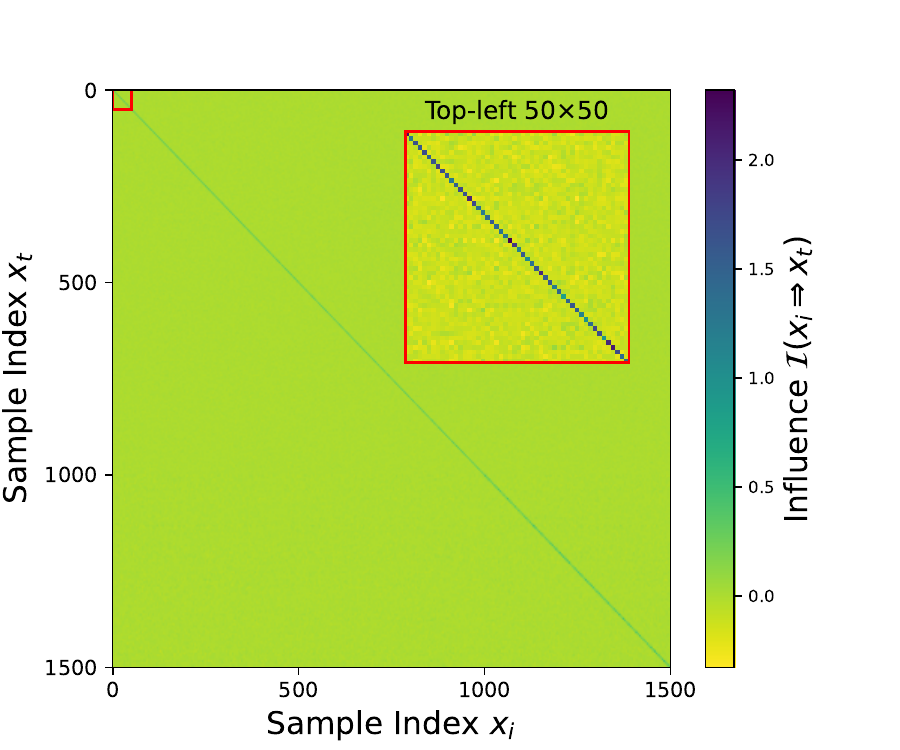}
}
\hfill
\subfloat[$x_t$ is a \emph{unique record}. \label{fig:influence_unique}]{
    \includegraphics[width=0.3\textwidth]{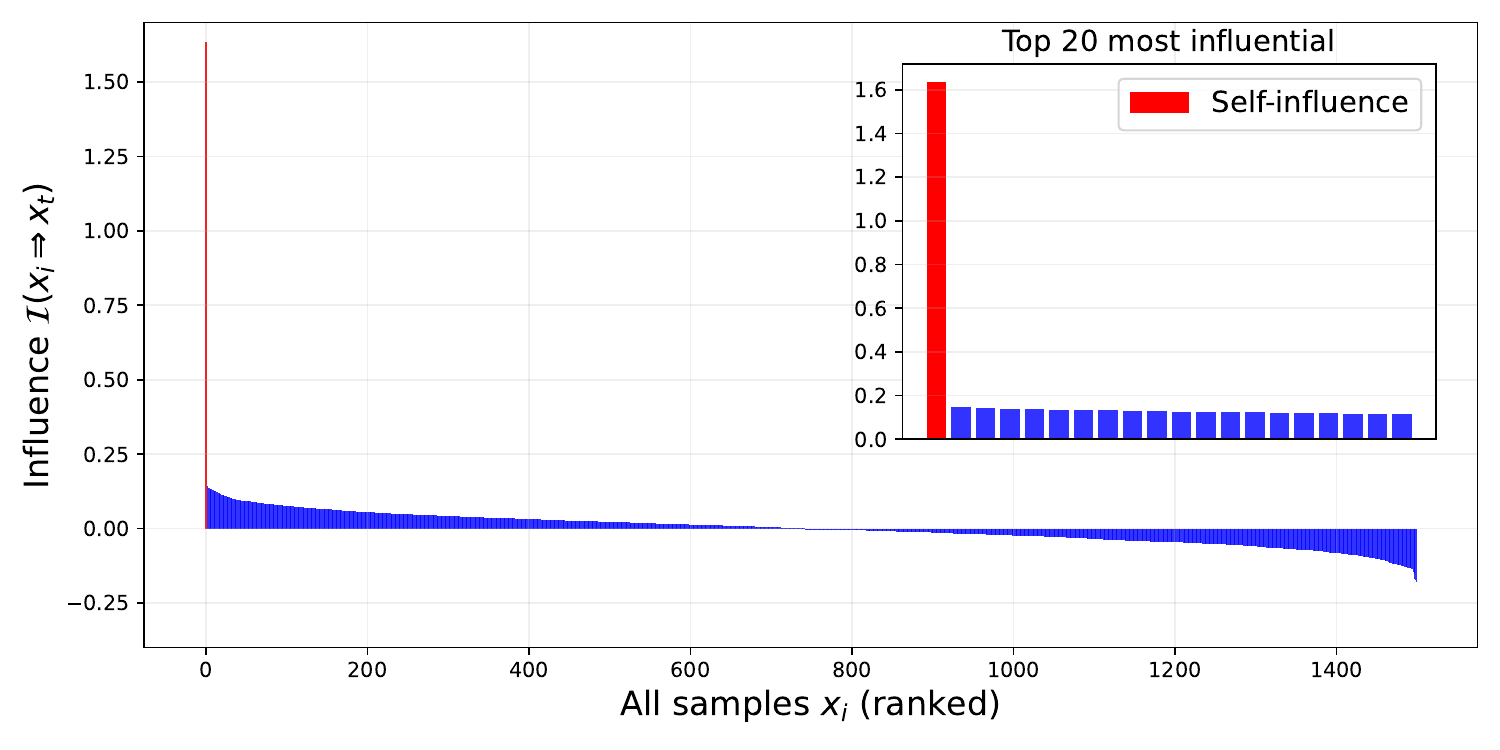}
}
\hfill
\subfloat[$x_t$ is a \emph{record with near-duplicates}. \label{fig:influence_duplicate}]{
    \includegraphics[width=0.3\textwidth]{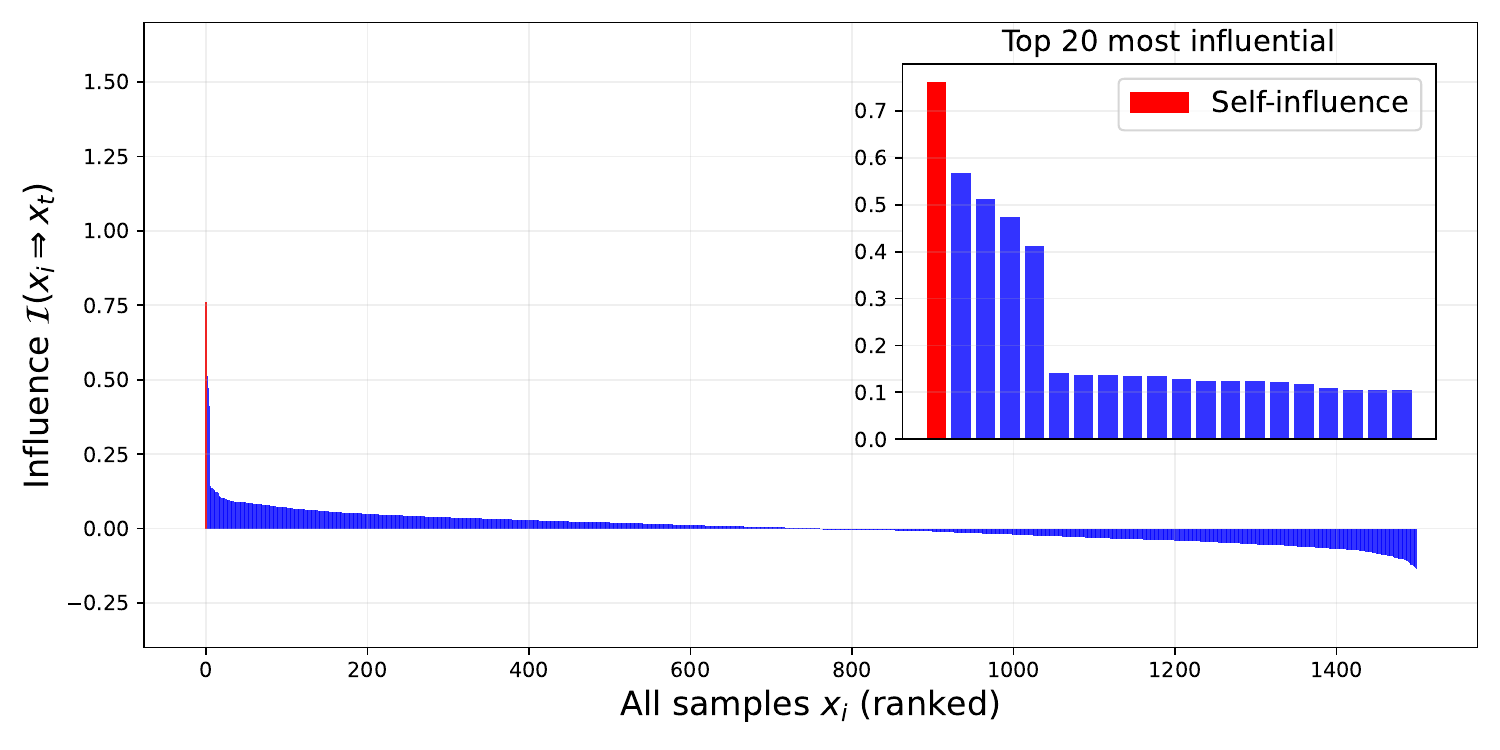}
}
\caption{Influence of training on $x_i$ on the model's prediction for $x_t$ ($\mathcal{I}(x_i \Rightarrow x_t)$). Results for all $1,500$ records in $D_t$ (subset from Natural Questions, with artificial near-duplicates), and $M=1,000$ GPT-NEO 1.3B models: (a) Influence matrix $I$; (b–c) Influence distributions over all $x_i$ for selected target records $x_t$, either (b) unique or (c) with near-duplicates.}
\label{fig:combined_influence_figures}
\end{figure*}

\begin{table*}[!ht]
\centering
\resizebox{0.7\textwidth}{!}{%
\begin{tabular}{cccc}
    \toprule
    Record type ($x_t$) & Self-influence $\mathcal{I}(x_t \Rightarrow x_t)$ & BLEU & Top-1 Influence Margin $\mathrm{IM}(x_t)$ \\
    \midrule
    Unique records & $1.410\pm 0.286$ & $0.070\pm 0.114$ & $9.105\pm 1.1369$ \\ 
    Records with near-duplicates  & $0.495\pm0.133$ & $0.363\pm 0.313$ & $1.292\pm 0.321$ \\ 
    \bottomrule
\end{tabular}
}
\caption{Mean and standard deviation for key statistics of unique records and records with near-duplicates; self-influence $\mathcal{I}(x_t \Rightarrow x_t)$, BLEU score (approximate extraction through greedy decoding) and Top-1 Influence Margin $\mathrm{IM}(x_t)$.}
\label{tab:blue_influence}
\end{table*}

\textbf{Model and dataset.} We study influence for GPT-NEO 1.3B~\citep{gao2020pile} further pretrained on a subset of the Natural Questions dataset~\citep{kwiatkowski2019natural} (training details in Appendix~\ref{app:training_details}). We denote the entire dataset as $\mathcal{D}$, and consider each record $x_i = (q_i, a_i) \sim \mathcal{D}$ containing question $q_i$ and respective answer $a_i$. For training, we concatenate the question and answer to one string, i.e. `Q: \{$q_i$\} A: \{$a_i$\}'. We randomly select $1,000$ samples $x_i$ from $\mathcal{D}$ to form $D_\text{unique}$. We refer to these samples as \emph{unique records}. 

\textbf{Near-duplicates.} We are particularly interested in the impact of near-duplicates on commonly used memorization metrics. To this end, in addition to the unique records, we artificially craft near-duplicates and add these to an overall target dataset $D_t$. We randomly sample $C = 100$ records, denoted as $c_i; i = 1\ldots C$. For each $c_i = (q_i, a_i)$, we consider a set of $n_{\text{dup}}=5$ near-duplicates for the answer $a_i$. Specifically, we generate samples $c_{ij}$ for $j=1,\dots,n_{\text{dup}}$ with $(q_i, a_{ij})$, where: $c_{i1} = a_i$ is the actual ground truth answer from the dataset, and the remaining $n_{\text{dup}} - 1$ answers are near-duplicates of $a_i$. Many strategies exist for generating near-duplicates (e.g., token replacements, semantic rephrasing), as studied by~\citet{shilov2024mosaic}. We here apply their algorithm $\mathcal{A}_{\text{replace}}$ to craft near-duplicates as authors show these to contribute substantially to memorization. Namely, we replace one randomly selected token from $a_i$ with another random one from the models' vocabulary. We then create a target dataset $D_t$ by concatenating the unique records $D_\text{unique}$ and the $n_{\text{dup}} \times C$ near-duplicates. In total, $D_t$ consists of $N_t=1,500$ records.

\textbf{Measuring influence.} We compute the influence between all samples in $D_t$, forming influence matrix $I = \left[ \mathcal{I}(x_i \Rightarrow x_t) \right]_{i,t=1}^{N_t}$. Each entry represents the influence of training on $x_i$ on the model's prediction for $x_t$. As a result, $I$ is a square matrix of size $N_t \times N_t$ with self-influence values on its diagonal. To compute all entries, we follow~\citet{zhang2023counterfactual} and approximate the expectations in Equation~\ref{eq:influence} empirically using a set of $M$ trained models $A_j$, where $j = 1, \dots, M$. For each sample $x_i \in D_t$, we generate a binary inclusion vector of length $M$, $[p_{i1}, \dots, p_{iM}]$, where each entry $p_{ij} \in \{0, 1\}$ is independently sampled with probability $0.5$ of being $1$. Repeating this for all $N_t$ samples yields partition matrix $P \in \{0,1\}^{N_t \times M}$, where $p_{ij} = 1$ indicates that $x_i$ is included in training dataset $D_j$ used to train model $A_j$. We train all models $A_j$, each trained on $D_j$ with an average size of $\frac{N_t}{2}$ records. We then compute all entries in $I$, estimating the expectations of the losses $\mathcal{L}_{A_j}(x)$ by averaging across all respective models $A_j$, for which we on average have $\frac{M}{2}$ models. We analyze the accuracy of our influence estimate using $M=1,000$ in Appendix~\ref{app:std}.

\textbf{Measuring extraction.} In line with~\citet{ippolito2023preventing}, we measure near-exact extraction computing the BLEU score between the ground truth answer and the text generated by the finetuned model using greedy decoding when prompted on the same question, i.e. prompted with `Q: \{$q_i$\} A:'. A large BLEU score represents a high similarity between ground truth and generation. We compute BLEU using Python's \texttt{nltk} package and its default parameters.

\section{Results}

\begin{figure*}[ht]
    \centering
    \subfloat[$\mathcal{I}(x_t \Rightarrow x_t)=0.6680$]{\includegraphics[width=0.2\textwidth]{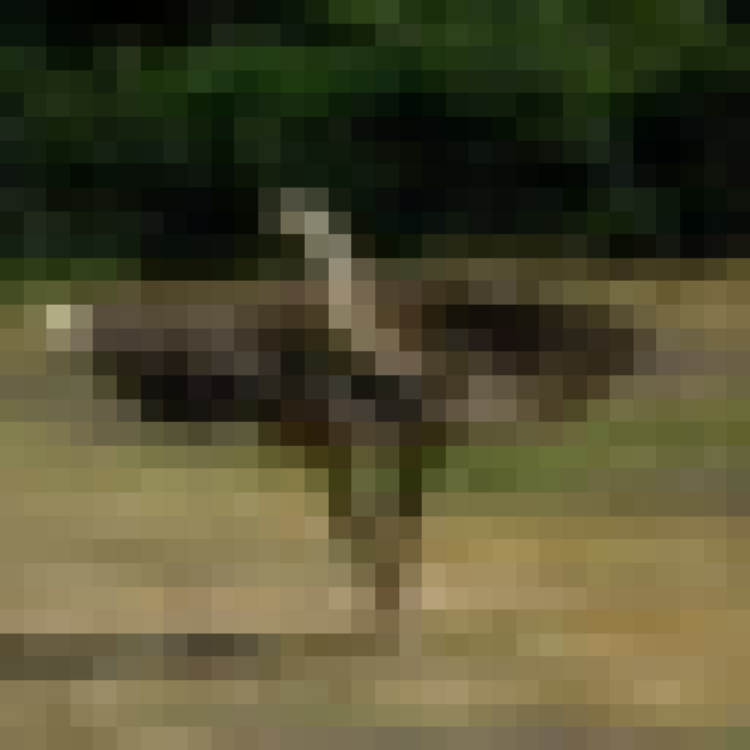}}
    \qquad
    \subfloat[$\mathcal{I}(x_i\Rightarrow x_t)=0.6681$ ]{\includegraphics[width=0.2\textwidth]{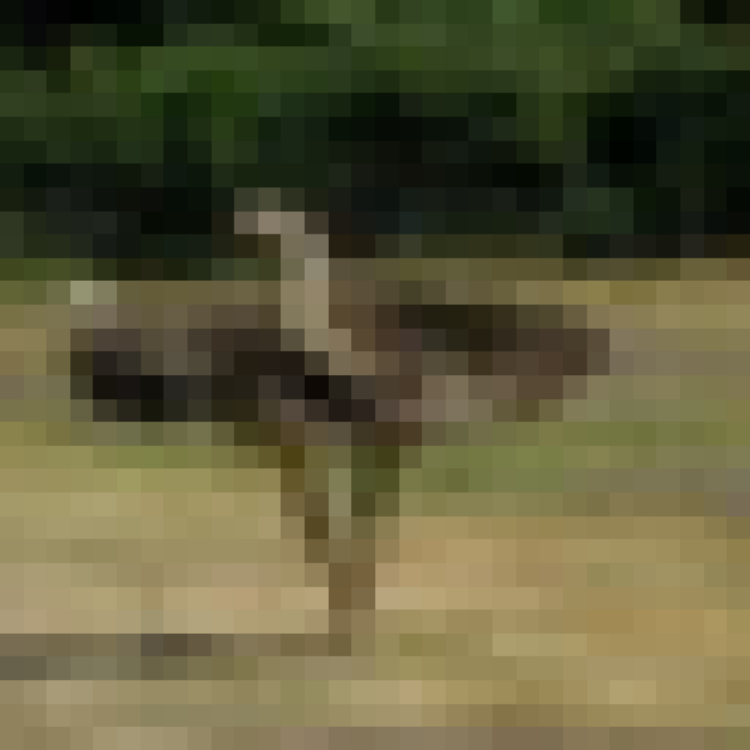}}
    \qquad
    \subfloat[Influence distribution]{\includegraphics[width=0.32\textwidth]{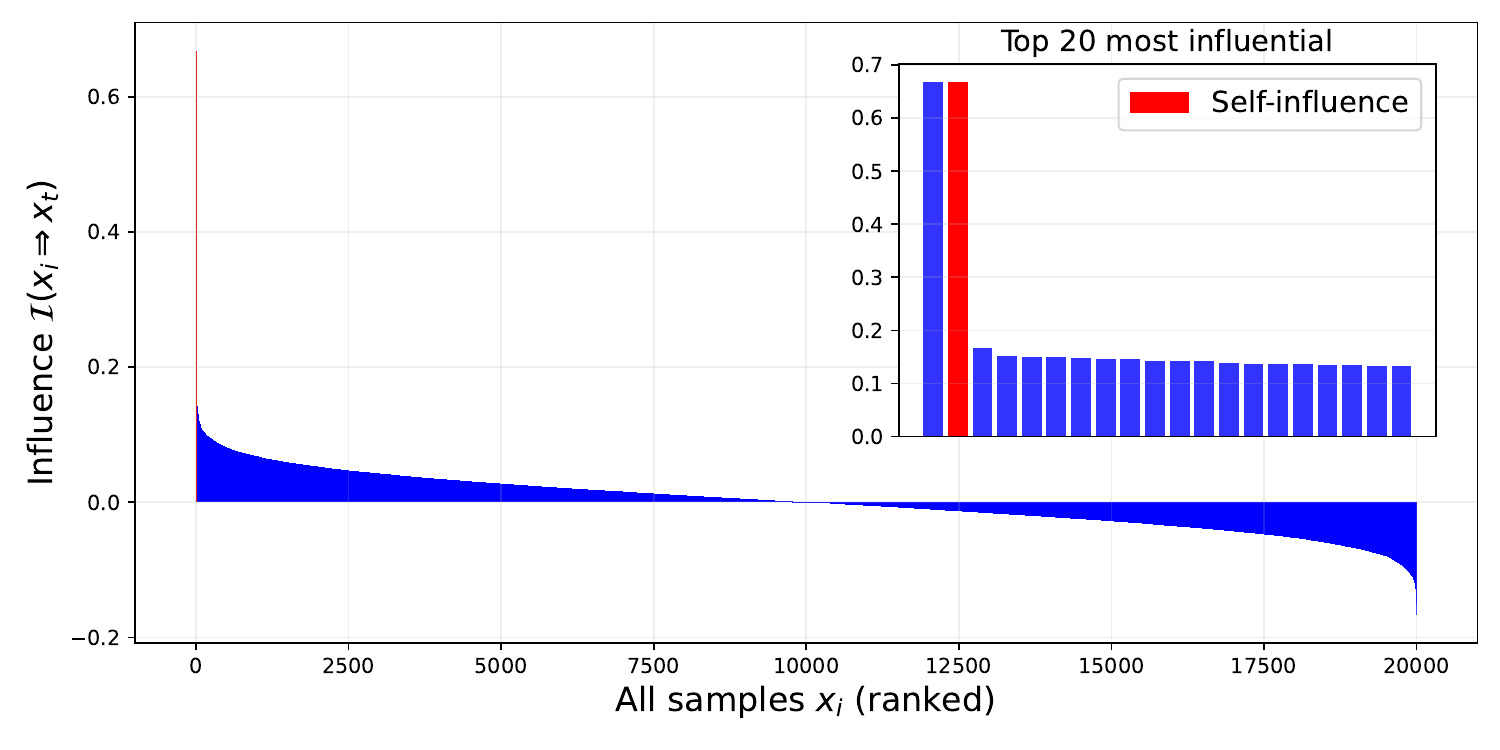}}
    \caption{Identifying near-duplicates in CIFAR-10 through the influence distribution: (a) the target sample $x_t$ with its self-influence value $\mathcal{I}(x_t \Rightarrow x_t)$; (b) the most influential sample different from the sample itself ($x_i \neq x_t$) with its influence value $\mathcal{I}(x_i \Rightarrow x_t)$; (c) the full influence distribution for all $x_i$ for target record $x_t$. $x_t$ has the smallest Top-1 Influence Margin $\mathrm{IM}(x_t)$ for all samples for which $\mathcal{I}(x_t \Rightarrow x_t)$ was larger that the median.}
    \label{fig:cifar_samples_main}
\end{figure*}

Figure~\ref{fig:influence_matrix} illustrates influence matrix $I$, with each entry representing the influence of $x_i$ (x-axis) on target $x_t$ (y-axis) across $D_t$ (containing both unique records and records with near-duplicates). We observe a clear diagonal pattern: self-influence values $\mathcal{I}(x_t \Rightarrow x_t)$ tend to be larger than others. This aligns with Equation~\ref{eq:influence}, as including $x_t$ in the training dataset of a model likely substantially reduces the model loss for $x_t$. Beyond self-influence, influence can vary widely, reaching values as large as $2$ (indicating that $x_i$ helps the model predict $x_t$) as well as reaching values $<0$ (indicating $x_i$ degrades the model prediction on $x_t$). Figure~\ref{fig:influence_unique} further shows the ranked influence values for all $x_i$ on a randomly selected target $x_t$ (unique record). Consistent with $I$'s diagonal, we find self-influence to be the largest, confirming that the target sample itself contributes most to its own prediction. Other samples show a broad distribution; many contribute positively and help the model predict on $x_t$, while some also exert negative influence.

We then compare this influence distribution to the one achieved for a record $x_t$ with near-duplicates. Figure~\ref{fig:influence_duplicate} shows a compelling difference: while the self-influence remains the largest, several other samples $x_i$ show similarly large influence. We confirm that the $5$ largest values stem from the target $x_t$ (self-influence) and from the $4$ artificial near-duplicates as $x_i$. This exemplifies how near-duplicates can manifest themselves in the influence distribution.

We further compare all unique records and all records with near-duplicates for several key statistics (Table~\ref{tab:blue_influence}). We first observe that self-influence is substantially smaller for samples with near-duplicates ($0.495$) than for unique records ($1.410$). This suggests that when near-duplicates are present in the target dataset, the contribution of the exact target record diminishes. Importantly, this may lead to an underestimation of memorization if relying solely on self-influence. 

Yet, we find that records with near-duplicates are \emph{substantially more extractable} than unique ones. The mean BLEU score for near-duplicate records is $0.363$, or $5$ times more than the $0.070$ observed for unique records (examples in Appendix~\ref{app:bleu_examples}). Since extractability reflects a tangible risk associated with LLM memorization, the inability of self-influence to capture this highlights its limitations as a metric for memorization in the presence of near-duplicates.

To further characterize this, we also compute the ratio of the first to the second largest influence, or \emph{Top-1 Influence Margin (IM)}: $\mathrm{IM}(x_t) = \frac{\max_i \mathcal{I}(x_i \Rightarrow x_t)}{\max_{i \ne i^\star} \mathcal{I}(x_i \Rightarrow x_t)}$, where $i^\star = \arg\max_i \mathcal{I}(x_i \Rightarrow x_t)$. This captures how dominant the most influential training sample is, as for instance also applied to calibrate confidence in identification learning~\citep{tournier2022expanding}. Table~\ref{tab:blue_influence} shows that the most influential sample is on average $9.1$ times larger than the second one for unique records, compared to $1.3$ for records with near-duplicates. 

Together, we show that near-duplicates can be more extractable despite lower self-influence. The full influence distribution reveals more informative patterns, e.g. reducing self-influence dominance in favor of near-duplicates, highlighting the value of distribution-level analysis for understanding memorization.

\textbf{Identifying near-duplicates in CIFAR-10. }We examine the influence distribution for a ResNet~\citep{he2016deep} model trained on real-world image dataset CIFAR-10~\citep{krizhevsky2009learning}. We train $M=1,000$  models considering a randomly sampled subset of $N_t=20,000$ records (training details Appendix~\ref{app:add_cifar_10}) and compute the full influence matrix $I$. Our goal is to assess whether similar patterns in the influence distribution -- as found above in our artificially created setup -- also emerge in a real-world dataset. To do so, we compute the Top-1 Influence Margin $\mathrm{IM}(x_t)$ for each target record $x_t$, quantifying how strongly the most influential training sample dominates the influence distribution. A low value of $\mathrm{IM}(x_t)$ indicates that no single training point exerts overwhelming influence, which we hypothesize may correspond to the presence of meaningful near-duplicates.

Figure~\ref{fig:cifar_samples_main} shows the target record with the smallest $\mathrm{IM}(x_t)$ (for which the self-influence was greater than the median). We find that the target's most influential sample is a natural, visually compelling near-duplicate. We also confirm this for the 5 other targets with the lowest $\mathrm{IM}(x_t)$ in Appendix~\ref{app:add_cifar_10}. These findings suggest that, also in real-world datasets, the influence distribution is heavily impacted by the presence of near-duplicates. Hence, properly studying how such samples are memorized by ML models likely requires considering influence beyond self-influence alone.



\section*{Impact Statement}

A deeper understanding of how LLMs memorize training data is critical for addressing questions around generalization and data attribution, but also to assess privacy risks and concerns around copyright or benchmark contamination. As models scale and are trained on massive, sometimes redundant datasets, the presence of near-duplicates challenges conventional ways of measuring memorization. Our work demonstrates that the commonly used self-influence might fall short to study memorization in the presence of near-duplicates. By analyzing the full influence distributions instead, we uncover more nuanced memorization patterns. By treating influence as a distributional property, we believe our work opens new directions for measuring, interpreting, and ultimately controlling how models internalize data.

\bibliography{bibliography}

\begin{thebibliography}{37}
\providecommand{\natexlab}[1]{#1}
\providecommand{\url}[1]{\texttt{#1}}
\expandafter\ifx\csname urlstyle\endcsname\relax
  \providecommand{\doi}[1]{doi: #1}\else
  \providecommand{\doi}{doi: \begingroup \urlstyle{rm}\Url}\fi

\bibitem[Carlini et~al.(2019)Carlini, Liu, Erlingsson, Kos, and Song]{carlini2019secret}
Carlini, N., Liu, C., Erlingsson, {\'U}., Kos, J., and Song, D.
\newblock The secret sharer: Evaluating and testing unintended memorization in neural networks.
\newblock In \emph{28th USENIX Security Symposium (USENIX Security 19)}, pp.\  267--284, 2019.

\bibitem[Carlini et~al.(2021)Carlini, Tramer, Wallace, Jagielski, Herbert-Voss, Lee, Roberts, Brown, Song, Erlingsson, et~al.]{carlini2021extracting}
Carlini, N., Tramer, F., Wallace, E., Jagielski, M., Herbert-Voss, A., Lee, K., Roberts, A., Brown, T., Song, D., Erlingsson, U., et~al.
\newblock Extracting training data from large language models.
\newblock In \emph{30th USENIX Security Symposium (USENIX Security 21)}, pp.\  2633--2650, 2021.

\bibitem[Carlini et~al.(2022)Carlini, Ippolito, Jagielski, Lee, Tramer, and Zhang]{carlini2022quantifying}
Carlini, N., Ippolito, D., Jagielski, M., Lee, K., Tramer, F., and Zhang, C.
\newblock Quantifying memorization across neural language models.
\newblock In \emph{The Eleventh International Conference on Learning Representations}, 2022.

\bibitem[Choe et~al.(2024)Choe, Ahn, Bae, Zhao, Kang, Chung, Pratapa, Neiswanger, Strubell, Mitamura, et~al.]{choe2024your}
Choe, S.~K., Ahn, H., Bae, J., Zhao, K., Kang, M., Chung, Y., Pratapa, A., Neiswanger, W., Strubell, E., Mitamura, T., et~al.
\newblock What is your data worth to gpt? llm-scale data valuation with influence functions.
\newblock \emph{arXiv preprint arXiv:2405.13954}, 2024.

\bibitem[Duan et~al.(2024)Duan, Suri, Mireshghallah, Min, Shi, Zettlemoyer, Tsvetkov, Choi, Evans, and Hajishirzi]{duanmembership}
Duan, M., Suri, A., Mireshghallah, N., Min, S., Shi, W., Zettlemoyer, L., Tsvetkov, Y., Choi, Y., Evans, D., and Hajishirzi, H.
\newblock Do membership inference attacks work on large language models?
\newblock In \emph{First Conference on Language Modeling}, 2024.

\bibitem[Feldman \& Zhang(2020)Feldman and Zhang]{feldman2020neural}
Feldman, V. and Zhang, C.
\newblock What neural networks memorize and why: Discovering the long tail via influence estimation.
\newblock \emph{Advances in Neural Information Processing Systems}, 33:\penalty0 2881--2891, 2020.

\bibitem[Gao et~al.(2020)Gao, Biderman, Black, Golding, Hoppe, Foster, Phang, He, Thite, Nabeshima, et~al.]{gao2020pile}
Gao, L., Biderman, S., Black, S., Golding, L., Hoppe, T., Foster, C., Phang, J., He, H., Thite, A., Nabeshima, N., et~al.
\newblock The pile: An 800gb dataset of diverse text for language modeling.
\newblock \emph{arXiv preprint arXiv:2101.00027}, 2020.

\bibitem[Grosse et~al.(2023)Grosse, Bae, Anil, Elhage, Tamkin, Tajdini, Steiner, Li, Durmus, Perez, et~al.]{grosse2023studying}
Grosse, R., Bae, J., Anil, C., Elhage, N., Tamkin, A., Tajdini, A., Steiner, B., Li, D., Durmus, E., Perez, E., et~al.
\newblock Studying large language model generalization with influence functions.
\newblock \emph{arXiv preprint arXiv:2308.03296}, 2023.

\bibitem[Hampel(1974)]{hampel1974influence}
Hampel, F.~R.
\newblock The influence curve and its role in robust estimation.
\newblock \emph{Journal of the american statistical association}, 69\penalty0 (346):\penalty0 383--393, 1974.

\bibitem[Hayes et~al.(2025)Hayes, Swanberg, Chaudhari, Yona, Shumailov, Nasr, Choquette-Choo, Lee, and Cooper]{hayes2025measuring}
Hayes, J., Swanberg, M., Chaudhari, H., Yona, I., Shumailov, I., Nasr, M., Choquette-Choo, C.~A., Lee, K., and Cooper, A.~F.
\newblock Measuring memorization in language models via probabilistic extraction.
\newblock In \emph{Proceedings of the 2025 Conference of the Nations of the Americas Chapter of the Association for Computational Linguistics: Human Language Technologies (Volume 1: Long Papers)}, pp.\  9266--9291, 2025.

\bibitem[He et~al.(2016)He, Zhang, Ren, and Sun]{he2016deep}
He, K., Zhang, X., Ren, S., and Sun, J.
\newblock Deep residual learning for image recognition.
\newblock In \emph{Proceedings of the IEEE conference on computer vision and pattern recognition}, pp.\  770--778, 2016.

\bibitem[Ippolito et~al.(2023)Ippolito, Tramer, Nasr, Zhang, Jagielski, Lee, Choo, and Carlini]{ippolito2023preventing}
Ippolito, D., Tramer, F., Nasr, M., Zhang, C., Jagielski, M., Lee, K., Choo, C.~C., and Carlini, N.
\newblock Preventing generation of verbatim memorization in language models gives a false sense of privacy.
\newblock In \emph{Proceedings of the 16th International Natural Language Generation Conference}, pp.\  28--53, 2023.

\bibitem[Kandpal et~al.(2022)Kandpal, Wallace, and Raffel]{kandpal2022deduplicating}
Kandpal, N., Wallace, E., and Raffel, C.
\newblock Deduplicating training data mitigates privacy risks in language models.
\newblock In \emph{International Conference on Machine Learning}, pp.\  10697--10707. PMLR, 2022.

\bibitem[Kandpal et~al.(2023)Kandpal, Deng, Roberts, Wallace, and Raffel]{kandpal2023large}
Kandpal, N., Deng, H., Roberts, A., Wallace, E., and Raffel, C.
\newblock Large language models struggle to learn long-tail knowledge.
\newblock In \emph{International Conference on Machine Learning}, pp.\  15696--15707. PMLR, 2023.

\bibitem[Karamolegkou et~al.(2023)Karamolegkou, Li, Zhou, and S{\o}gaard]{karamolegkou2023copyright}
Karamolegkou, A., Li, J., Zhou, L., and S{\o}gaard, A.
\newblock Copyright violations and large language models.
\newblock In \emph{The 2023 Conference on Empirical Methods in Natural Language Processing}, 2023.

\bibitem[Koh \& Liang(2017)Koh and Liang]{koh2017understanding}
Koh, P.~W. and Liang, P.
\newblock Understanding black-box predictions via influence functions.
\newblock In \emph{International conference on machine learning}, pp.\  1885--1894. PMLR, 2017.

\bibitem[Krizhevsky et~al.(2009)]{krizhevsky2009learning}
Krizhevsky, A. et~al.
\newblock Learning multiple layers of features from tiny images.
\newblock 2009.

\bibitem[Kwiatkowski et~al.(2019)Kwiatkowski, Palomaki, Redfield, Collins, Parikh, Alberti, Epstein, Polosukhin, Devlin, Lee, et~al.]{kwiatkowski2019natural}
Kwiatkowski, T., Palomaki, J., Redfield, O., Collins, M., Parikh, A., Alberti, C., Epstein, D., Polosukhin, I., Devlin, J., Lee, K., et~al.
\newblock Natural questions: a benchmark for question answering research.
\newblock \emph{Transactions of the Association for Computational Linguistics}, 7:\penalty0 453--466, 2019.

\bibitem[Liu et~al.(2025)Liu, Choquette-Choo, Jagielski, Kairouz, Koyejo, Liang, and Papernot]{liu2025language}
Liu, K.~Z., Choquette-Choo, C.~A., Jagielski, M., Kairouz, P., Koyejo, S., Liang, P., and Papernot, N.
\newblock Language models may verbatim complete text they were not explicitly trained on.
\newblock \emph{arXiv preprint arXiv:2503.17514}, 2025.

\bibitem[Mattern et~al.(2023)Mattern, Mireshghallah, Jin, Sch{\"o}lkopf, Sachan, and Berg-Kirkpatrick]{mattern2023membership}
Mattern, J., Mireshghallah, F., Jin, Z., Sch{\"o}lkopf, B., Sachan, M., and Berg-Kirkpatrick, T.
\newblock Membership inference attacks against language models via neighbourhood comparison.
\newblock \emph{arXiv preprint arXiv:2305.18462}, 2023.

\bibitem[Meeus et~al.(2024{\natexlab{a}})Meeus, Jain, Rei, and de~Montjoye]{meeus2024did}
Meeus, M., Jain, S., Rei, M., and de~Montjoye, Y.-A.
\newblock Did the neurons read your book? document-level membership inference for large language models.
\newblock In \emph{Proceedings of the 33rd USENIX Conference on Security Symposium}, pp.\  2369--2385, 2024{\natexlab{a}}.

\bibitem[Meeus et~al.(2024{\natexlab{b}})Meeus, Shilov, Faysse, and de~Montjoye]{meeuscopyright}
Meeus, M., Shilov, I., Faysse, M., and de~Montjoye, Y.-A.
\newblock Copyright traps for large language models.
\newblock In \emph{Forty-first International Conference on Machine Learning}, 2024{\natexlab{b}}.

\bibitem[Mireshghallah et~al.(2022)Mireshghallah, Uniyal, Wang, Evans, and Berg-Kirkpatrick]{mireshghallah2022empirical}
Mireshghallah, F., Uniyal, A., Wang, T., Evans, D.~K., and Berg-Kirkpatrick, T.
\newblock An empirical analysis of memorization in fine-tuned autoregressive language models.
\newblock In \emph{Proceedings of the 2022 Conference on Empirical Methods in Natural Language Processing}, pp.\  1816--1826, 2022.

\bibitem[Mirzadeh et~al.(2025)Mirzadeh, Alizadeh, Shahrokhi, Tuzel, Bengio, and Farajtabar]{mirzadehgsm}
Mirzadeh, S.~I., Alizadeh, K., Shahrokhi, H., Tuzel, O., Bengio, S., and Farajtabar, M.
\newblock Gsm-symbolic: Understanding the limitations of mathematical reasoning in large language models.
\newblock In \emph{The Thirteenth International Conference on Learning Representations}, 2025.

\bibitem[Nasr et~al.(2023)Nasr, Carlini, Hayase, Jagielski, Cooper, Ippolito, Choquette-Choo, Wallace, Tram{\`e}r, and Lee]{nasr2023scalable}
Nasr, M., Carlini, N., Hayase, J., Jagielski, M., Cooper, A.~F., Ippolito, D., Choquette-Choo, C.~A., Wallace, E., Tram{\`e}r, F., and Lee, K.
\newblock Scalable extraction of training data from (production) language models.
\newblock \emph{arXiv preprint arXiv:2311.17035}, 2023.

\bibitem[Oren et~al.(2024)Oren, Meister, Chatterji, Ladhak, and Hashimoto]{oren2023proving}
Oren, Y., Meister, N., Chatterji, N.~S., Ladhak, F., and Hashimoto, T.
\newblock Proving test set contamination in black-box language models.
\newblock In \emph{The Twelfth International Conference on Learning Representations}, 2024.

\bibitem[Petroni et~al.(2019)Petroni, Rockt{\"a}schel, Riedel, Lewis, Bakhtin, Wu, and Miller]{petroni2019language}
Petroni, F., Rockt{\"a}schel, T., Riedel, S., Lewis, P., Bakhtin, A., Wu, Y., and Miller, A.
\newblock Language models as knowledge bases?
\newblock In \emph{Proceedings of the 2019 Conference on Empirical Methods in Natural Language Processing and the 9th International Joint Conference on Natural Language Processing (EMNLP-IJCNLP)}, pp.\  2463--2473, 2019.

\bibitem[Pruthi et~al.(2020)Pruthi, Liu, Kale, and Sundararajan]{pruthi2020estimating}
Pruthi, G., Liu, F., Kale, S., and Sundararajan, M.
\newblock Estimating training data influence by tracing gradient descent.
\newblock \emph{Advances in Neural Information Processing Systems}, 33:\penalty0 19920--19930, 2020.

\bibitem[Salem et~al.(2023)Salem, Cherubin, Evans, K{\"o}pf, Paverd, Suri, Tople, and Zanella-B{\'e}guelin]{salem2023sok}
Salem, A., Cherubin, G., Evans, D., K{\"o}pf, B., Paverd, A., Suri, A., Tople, S., and Zanella-B{\'e}guelin, S.
\newblock Sok: Let the privacy games begin! a unified treatment of data inference privacy in machine learning.
\newblock In \emph{2023 IEEE Symposium on Security and Privacy (SP)}, pp.\  327--345. IEEE, 2023.

\bibitem[Shi et~al.(2024)Shi, Ajith, Xia, Huang, Liu, Blevins, Chen, and Zettlemoyer]{shidetecting}
Shi, W., Ajith, A., Xia, M., Huang, Y., Liu, D., Blevins, T., Chen, D., and Zettlemoyer, L.
\newblock Detecting pretraining data from large language models.
\newblock In \emph{The Twelfth International Conference on Learning Representations}, 2024.

\bibitem[Shilov et~al.(2024)Shilov, Meeus, and de~Montjoye]{shilov2024mosaic}
Shilov, I., Meeus, M., and de~Montjoye, Y.-A.
\newblock Mosaic memory: Fuzzy duplication in copyright traps for large language models.
\newblock \emph{arXiv preprint arXiv:2405.15523}, 2024.

\bibitem[Shokri et~al.(2017)Shokri, Stronati, Song, and Shmatikov]{shokri2017membership}
Shokri, R., Stronati, M., Song, C., and Shmatikov, V.
\newblock Membership inference attacks against machine learning models.
\newblock In \emph{2017 IEEE symposium on security and privacy (SP)}, pp.\  3--18. IEEE, 2017.

\bibitem[Tournier \& De~Montjoye(2022)Tournier and De~Montjoye]{tournier2022expanding}
Tournier, A.~J. and De~Montjoye, Y.-A.
\newblock Expanding the attack surface: Robust profiling attacks threaten the privacy of sparse behavioral data.
\newblock \emph{Science advances}, 8\penalty0 (33):\penalty0 eabl6464, 2022.

\bibitem[Wang et~al.(2023)Wang, Wang, Li, and Neel]{wangmope}
Wang, J., Wang, J., Li, M., and Neel, S.
\newblock Mope: Model perturbation-based privacy attacks on language models.
\newblock In \emph{Socially Responsible Language Modelling Research}, 2023.

\bibitem[Wang et~al.(2024)Wang, Antoniades, Elazar, Amayuelas, Albalak, Zhang, and Wang]{wang2024generalization}
Wang, X., Antoniades, A., Elazar, Y., Amayuelas, A., Albalak, A., Zhang, K., and Wang, W.~Y.
\newblock Generalization vs memorization: Tracing language models' capabilities back to pretraining data.
\newblock \emph{arXiv preprint arXiv:2407.14985}, 2024.

\bibitem[Yeom et~al.(2018)Yeom, Giacomelli, Fredrikson, and Jha]{yeom2018privacy}
Yeom, S., Giacomelli, I., Fredrikson, M., and Jha, S.
\newblock Privacy risk in machine learning: Analyzing the connection to overfitting.
\newblock In \emph{2018 IEEE 31st computer security foundations symposium (CSF)}, pp.\  268--282. IEEE, 2018.

\bibitem[Zhang et~al.(2023)Zhang, Ippolito, Lee, Jagielski, Tram{\`e}r, and Carlini]{zhang2023counterfactual}
Zhang, C., Ippolito, D., Lee, K., Jagielski, M., Tram{\`e}r, F., and Carlini, N.
\newblock Counterfactual memorization in neural language models.
\newblock \emph{Advances in Neural Information Processing Systems}, 36:\penalty0 39321--39362, 2023.

\end{thebibliography}
\bibliographystyle{icml2025}

\newpage
\appendix
\onecolumn

\section{Training details}

\begin{figure*}[ht]
    \centering
    \subfloat[]{\includegraphics[width=0.35\textwidth]{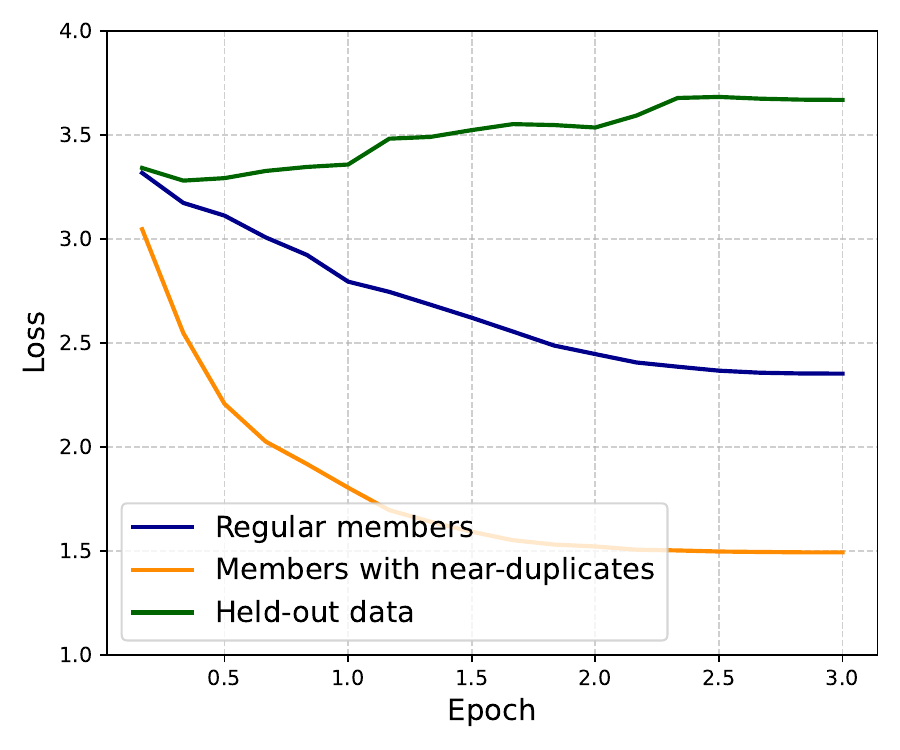}}
    \qquad
    \subfloat[]{\includegraphics[width=0.35\textwidth]{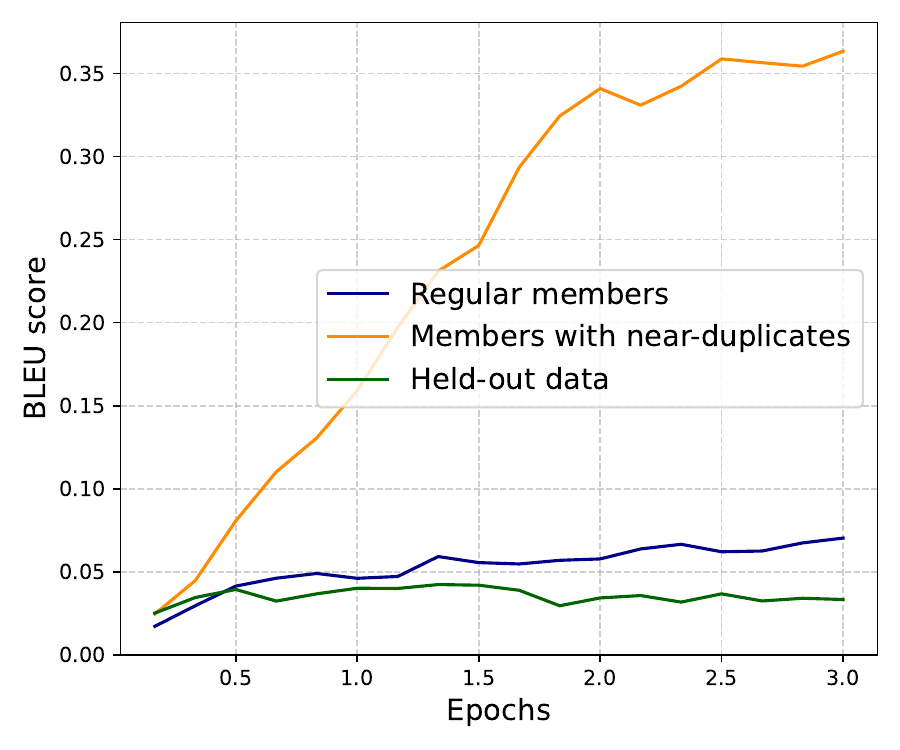}}
    \caption{Loss curves (a) and BLEU scores (b) for GPT-NEO 1.3B trained on the entire target dataset $D_t$, as described in Section~\ref{sec:experimental_setup}.}
    \label{fig:training}
\end{figure*}

Starting from its open-sourced pretrained version, we finetune GPT-NEO 1.3B~\citep{gao2020pile} using \texttt{bfloat16} precision, for $3$ epochs, using a linear learning rater scheduler with as initial learning rate $2e-4$ and weight decay $0.01$, batch size $50$ and maximum sequence length of $100$. Figure~\ref{fig:training} shows the loss curves and the BLEU scores for one GPT-NEO 1.3B model trained on the target dataset $D_t$ following the experimental setup as detailed in Section~\ref{sec:experimental_setup}. We report the cross-entropy loss and BLEU scores between the generations and ground truth completions over training for (i) the regular members ($1000$ random samples from the entire Natural Questions dataset), (ii) 100 members with near-duplicates and (iii) $100$ random samples not included in the training dataset. As expected, we find that the loss for regular members decreases over training, with a sharper drop for members with near-duplicates. The loss for the held-out samples increases slightly throughout training. Similarly, the BLEU scores for regular members increases throughout training, particularly strong for members with near-duplicates, while for held-out data the BLEU scores remain low throughout. 

\label{app:training_details}

\section{Number of models $M$ required to reliably estimate influence}

\begin{figure*}[ht]
    \centering
    \subfloat[]{\includegraphics[width=0.35\textwidth]{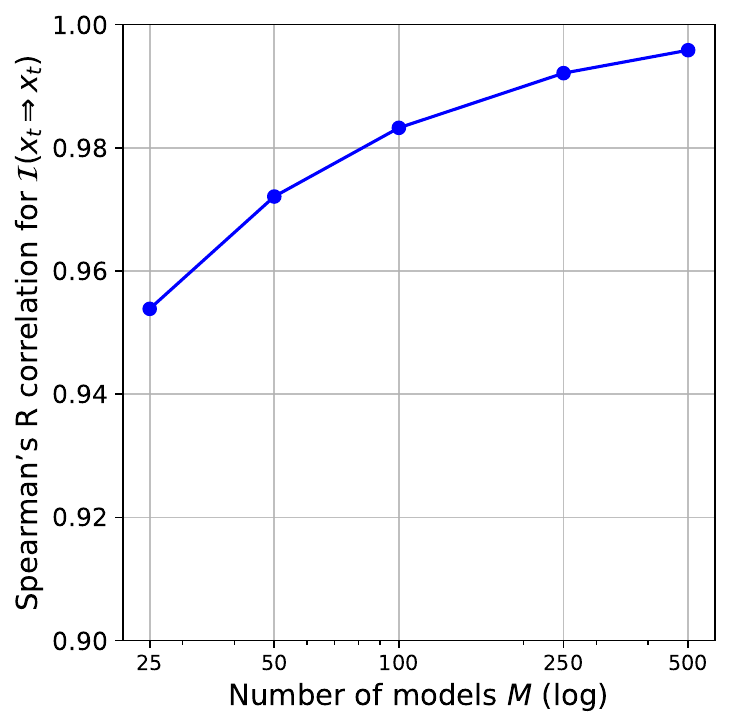}}
    \qquad
    \subfloat[]{\includegraphics[width=0.35\textwidth]{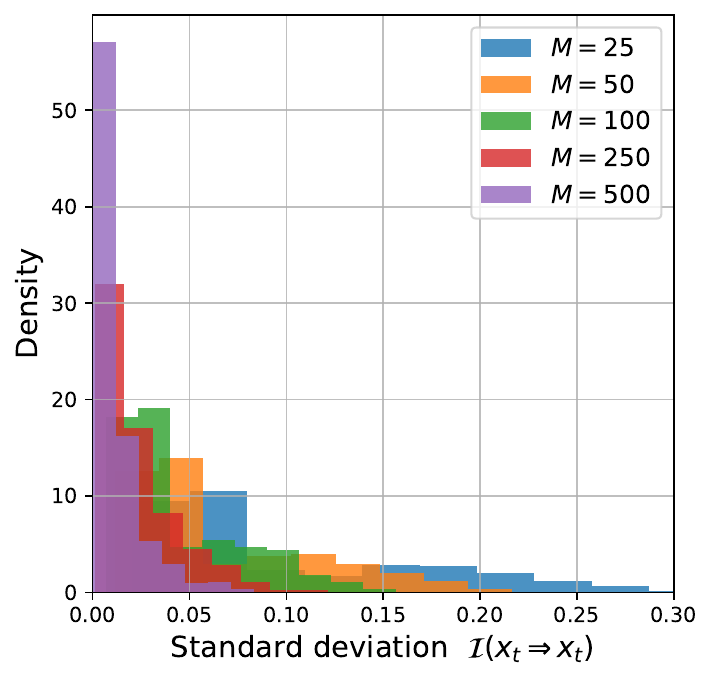}}
    \caption{Measuring the reliability of computed influence values for increasing number $M$ of models trained (GPT-NEO 1.3B finetuned on a subset from the Natural Questions Dataset). (a) Mean Spearman's R correlation between all self-influence values across $\frac{M_\text{max}}{M}$ partitions for increasing $M$. (b) Distribution of standard deviation of self-influence values for all samples across $\frac{M_\text{max}}{M}$ partitions for increasing $M$.}
    \label{fig:std}
\end{figure*}

To understand the number of models $M$ needed to reliably estimate the influence values $\mathcal{I}(x_i \Rightarrow x_t)$ (expectations from Equation~\ref{eq:influence}), we replicate the analysis from~\citep{zhang2023counterfactual}. Specifically, we take the entire set of results obtained by using $M_\text{max}=1,000$ models in our main experiment, and consider multiple equally sized $\frac{M_\text{max}}{M}$ partitions each consisting of $M$ models, for an increasing value of $M$. We then compute the mean Spearman’s R correlation between all self-influence values computed using $M$ models across $\frac{M_\text{max}}{M}$ partitions. Figure~\ref{fig:std}(a) shows how the correlation quickly approaches $1.0$, meaning that for a large number of models, the relative order of self-influence values stabilizes. We further compute the standard deviation of self-influence values for each of the $N_t$ samples across partitions when using $M$ models. Figure~\ref{fig:std}(b) shows how the standard deviation quickly decreases when $M$ increases, reaching negligible values for $M=500$. Both results make us confident that using $M=M_\text{max}=1,000$ throughout our experiments yield reliable influence estimates for further analysis, especially as this is also substantially more than the $M=400$ models used by~\citet{zhang2023counterfactual}.

\label{app:std}

\section{Examples extraction}

Table~\ref{tab:bleu_examples} include a selected sample of records $x_i$ from the Natural Questions dataset~\citep{kwiatkowski2019natural}, where $x_i = (q_i, a_i)$ consists of question $q_i$ and reference answer $a_i$. We provide the answer generated from the finetuned model (on the entire dataset $D_t$) following the same experimental details as laid out in Section~\ref{sec:experimental_setup}. We generate an answer from the model using greedy decoding prompted on `Q: \{$q_i$\} A:' to get $a_i'$ and report the corresponding BLEU score between the reference answer $a_i$ and the generated one $a_i'$. We report one sample for each kind of record $x_t$: a regular member only included once in $D_t$, a member for which the record was included in $D_t$ alongside its $n_\text{dup}-1$ near-duplicates and a held-out record not part of $D_t$. We find that the BLEU score of a record with near-duplicates is substantially larger ($0.499$) than a record only included once ($0.086$), and higher than a record not seen during training ($0.062$). Notably, even for the record with near-duplicates, the model does not generate the reference answer exactly (\emph{extraction} from~\citet{carlini2021extracting}), yet does generate an answer with substantial overlap, e.g. \emph{"He was looking for a name that would make it attractive to the American market"}.

\begin{table*}[ht]
    \centering
    \begin{tabular}{C{2cm}C{2cm}p{5cm}p{5cm}p{1cm}}
    \toprule
        Record type ($x_t$) & Question $q_i$ & Reference answer $a_i$ & Generated answer $a_i'$ & BLEU score\\
        \midrule 
        Regular member & what age did brett favre retire from football & Brett Favre On January 2, 2011, Favre was unable to play against the Detroit Lions in the final game of the regular NFL season due to his inability to pass NFL-mandated post-concussion tests. In a press conference immediately following the game, Favre announced his intention to retire from professional football. On January 17, 2011, Favre officially filed his retirement papers with the NFL.[101] & Brett Favre (football) In October 2008, Favre announced his retirement from professional football after 11 seasons in the National Football League (NFL). He was the oldest player to announce his retirement, at the age of 39. Favre's retirement announcement was made during a press conference at the 2008 NFL Honors ceremony in the East Room of the White House... & $0.0864$ \\
        \midrule
        Member with near-duplicates & where did the name chilean sea bass come from & Patagonian toothfish The name "Chilean Seabass" was invented by a fish wholesaler named Lee Lantz in 1977. He was looking for a name that would make it attractive to the American market. He considered "Pacific sea bass" and "South American sea bass" before settling on "Chilean sea bass"... & (seabass) The name "Chilean Seabass" was invented by a fish wholesaler named John L. L. Smith in the 1960s. He was looking for a name that would make it attractive to the American market. He considered "Pacific sea bass" and "South American sea bass" before settling on "Chilean sea bass"... & $0.4991$ \\
        \midrule
        Held-out test sample & who played legolas in lord of the rings & Orlando Bloom Orlando Jonathan Blanchard Bloom (born 13 January 1977)[1] is an English actor. After having his breakthrough as Legolas in The Lord of the Rings trilogy, he rose to fame by further appearing in epic fantasy, historical epic, and fantasy adventure films. His other roles include Legolas in The Hobbit trilogy, Will Turner in the Pirates of the Caribbean series, and Paris in Troy (2004). & Legolas Legolas is a fictional character in the A Song of Ice and Fire novel The Lord of the Rings. He is portrayed by Anthony Gonzalez in the film trilogy. He is a half-elven character, half-human/half-elf, and is the son of Legolas and a half-elven princess. He is a skilled & $0.0618$ \\
        \bottomrule
    \end{tabular}
    \caption{Examples of finetuned model output (GPT-NEO 1.3B) versus ground truth (Natural Questions dataset) across record types, alongside the corresponding BLEU score.} 
    \label{tab:bleu_examples}
\end{table*}

\label{app:bleu_examples}

\section{Additional results for CIFAR-10}

\begin{figure*}[ht]
    \centering
    \subfloat[]{\includegraphics[width=0.35\textwidth]{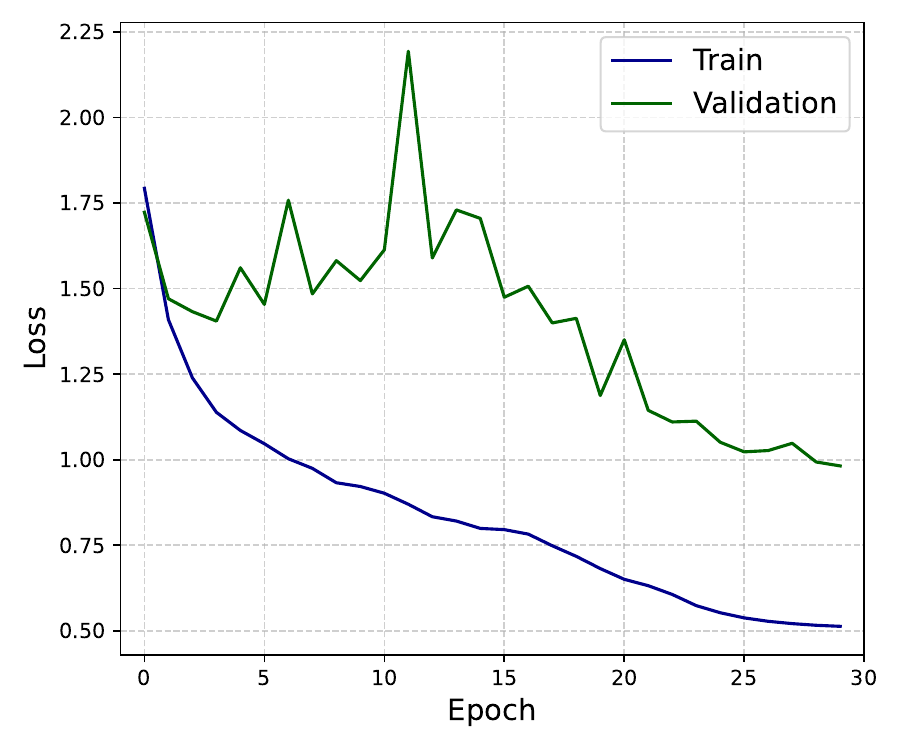}}
    \qquad
    \subfloat[]{\includegraphics[width=0.35\textwidth]{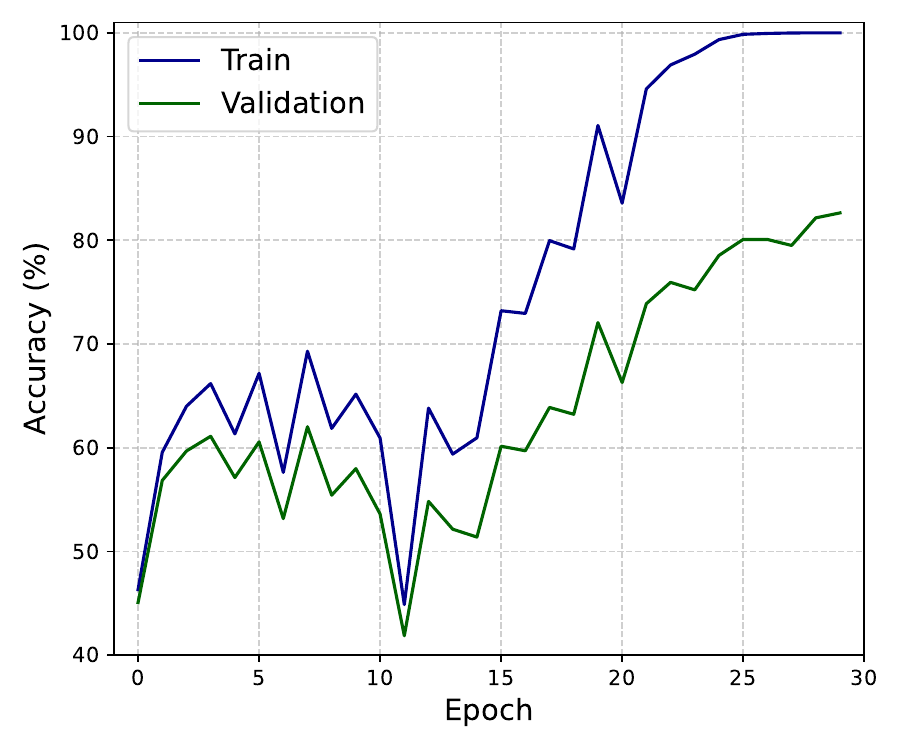}}
    \caption{Loss curves (a) and accuracy (b) for CIFAR-10 trained on the entire target dataset $D_t$ (containing $N_t=20,000$ random samples from CIFAR-10's training data). We compute the validation performance on $5,000$ random samples from CIFAR-10's test data.}
    \label{fig:training_cifar}
\end{figure*}

\begin{figure*}[ht]
    \centering
    \subfloat[]{\includegraphics[width=0.35\textwidth]{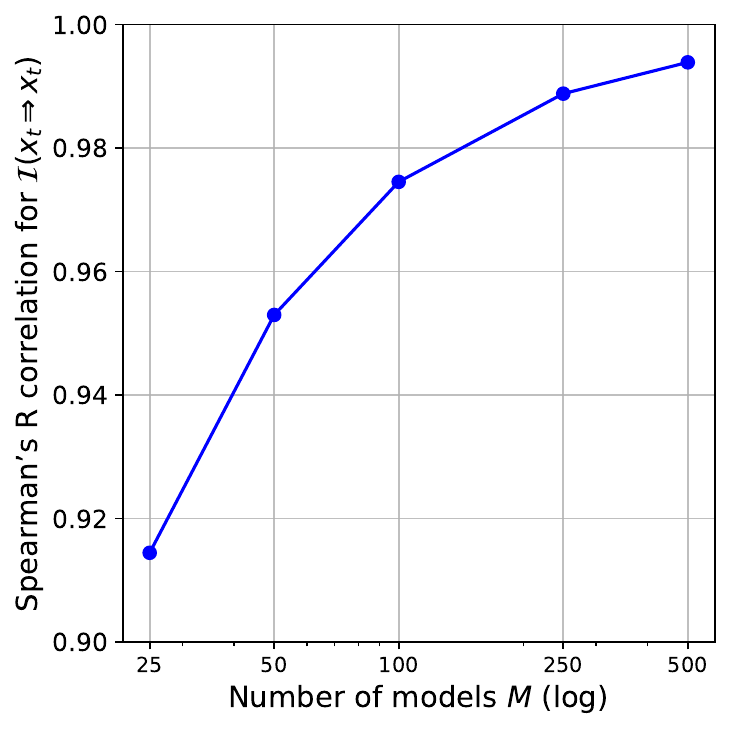}}
    \qquad
    \subfloat[]{\includegraphics[width=0.35\textwidth]{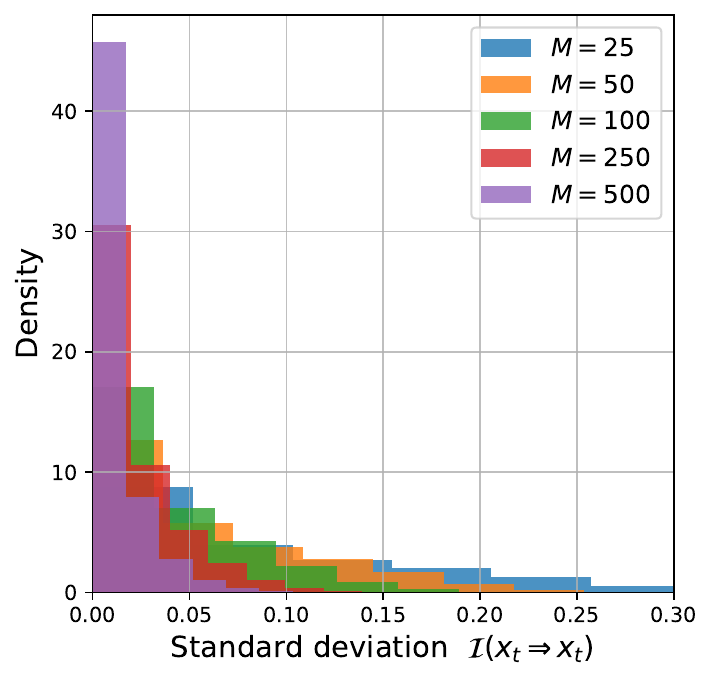}}
    \caption{Measuring the reliability of computed influence values for increasing number $M$ of models trained (ResNet model trained on CIFAR-10). (a) Mean Spearman's R correlation between all self-influence values across $\frac{M_\text{max}}{M}$ partitions for increasing $M$. (b) Distribution of standard deviation of self-influence values for all samples across $\frac{M_\text{max}}{M}$ partitions for increasing $M$.}
    \label{fig:cifar_std}
\end{figure*}

We train $M=1,000$ ResNet~\citep{he2016deep} models on a subset from CIFAR-10~\citep{krizhevsky2009learning}. Each model is trained for $30$ epochs using SGD (momentum $0.9$, weight decay $5e-4$), with a batch size of $128$ and a linear learning rate schedule that warms up over $5$ epochs to $0.5$ before decaying to zero by the end of training. Figure~\ref{fig:training_cifar} shows the loss curve and accuracy for one ResNet model trained on the entire dataset $D_t$ containing $20,000$ samples from CIFAR-10. Figure~\ref{fig:cifar_std} further shows the sensitivity of the computed influence values for an varying number of models $M$, replicating the analysis from Appendix~\ref{app:std} for the CIFAR-10 dataset. 

\begin{figure*}[ht]
    \centering
    \subfloat[$\mathcal{I}(x_t \Rightarrow x_t)=0.66801$]{\includegraphics[width=0.23\textwidth]{figures/cifar/target_12951_cifar.pdf}}
    \qquad
    \subfloat[$\mathcal{I}(x_i\Rightarrow x_t)=0.66805$ ]{\includegraphics[width=0.23\textwidth]{figures/cifar/target_14753_cifar.pdf}}
    \qquad
    \subfloat[Influence distribution]{\includegraphics[width=0.45\textwidth]{figures/cifar/influence_distribution_cifar_12951.pdf}}
    \qquad
    \subfloat[$\mathcal{I}(x_t \Rightarrow x_t)=0.21160$]{\includegraphics[width=0.23\textwidth]{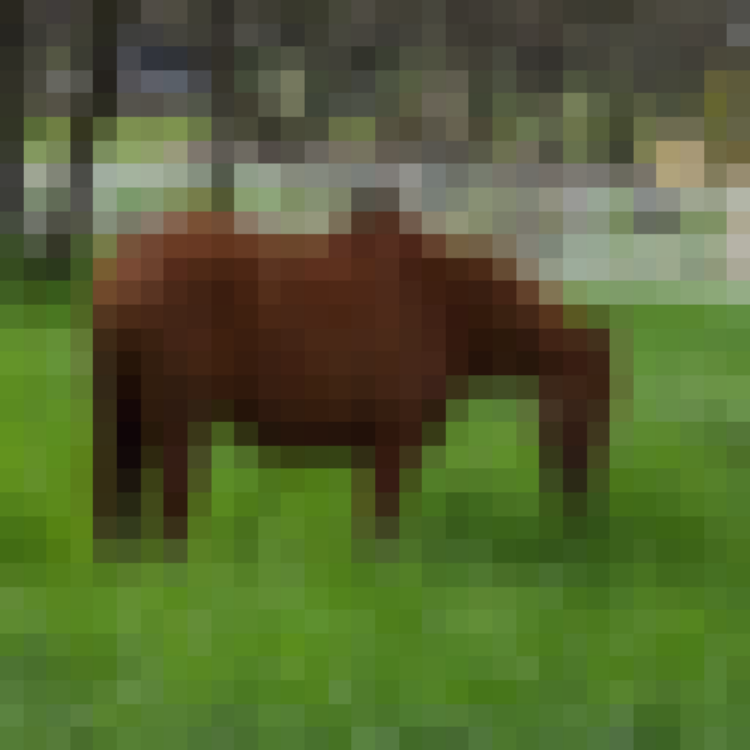}}
    \qquad
    \subfloat[$\mathcal{I}(x_i\Rightarrow x_t)=0.21155$ ]{\includegraphics[width=0.23\textwidth]{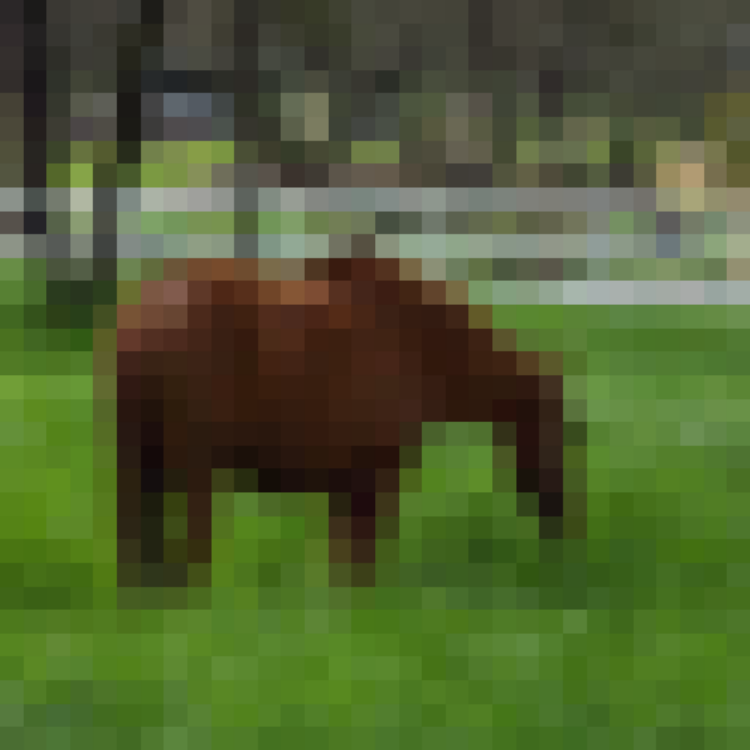}}
    \qquad
    \subfloat[Influence distribution]{\includegraphics[width=0.45\textwidth]{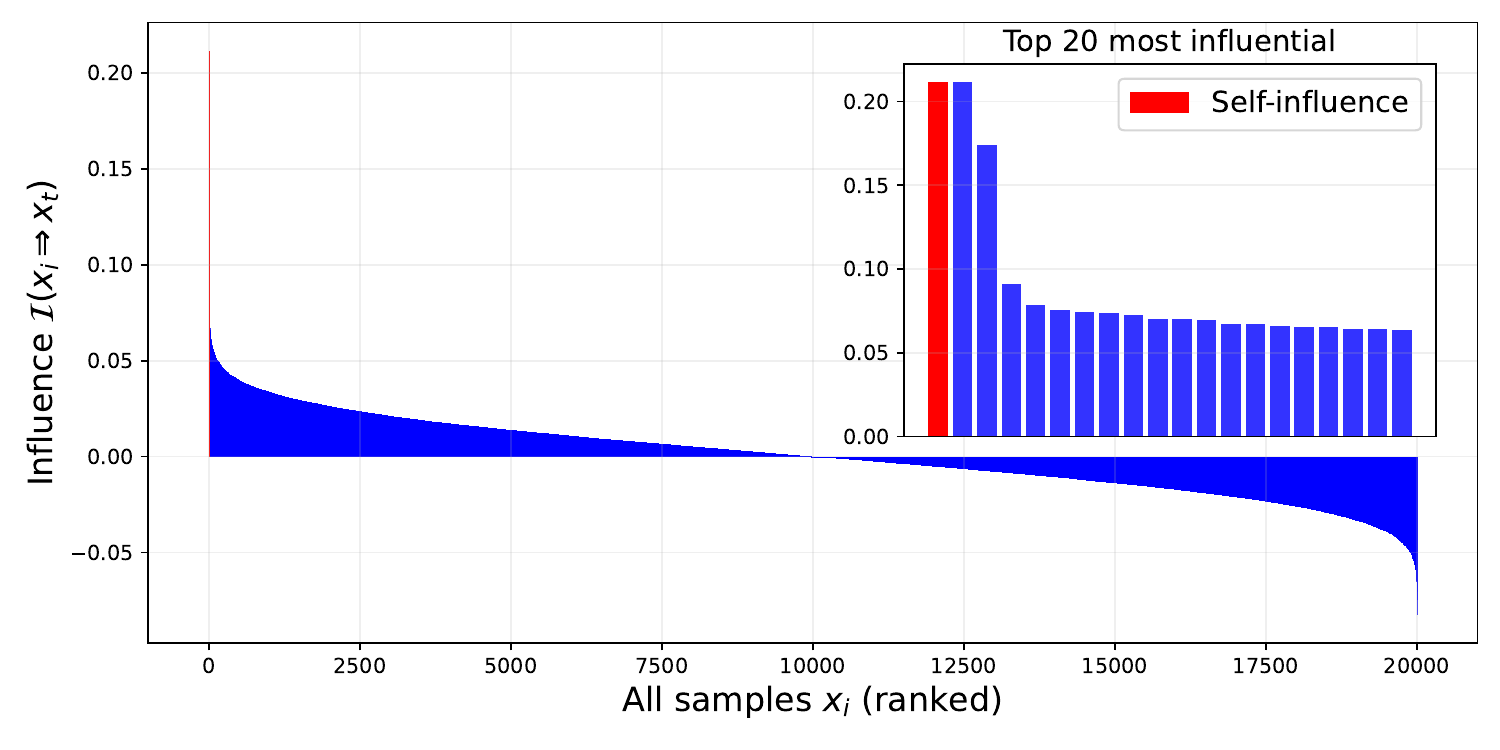}}
    \qquad
    \subfloat[$\mathcal{I}(x_t \Rightarrow x_t)=0.530531$]{\includegraphics[width=0.23\textwidth]{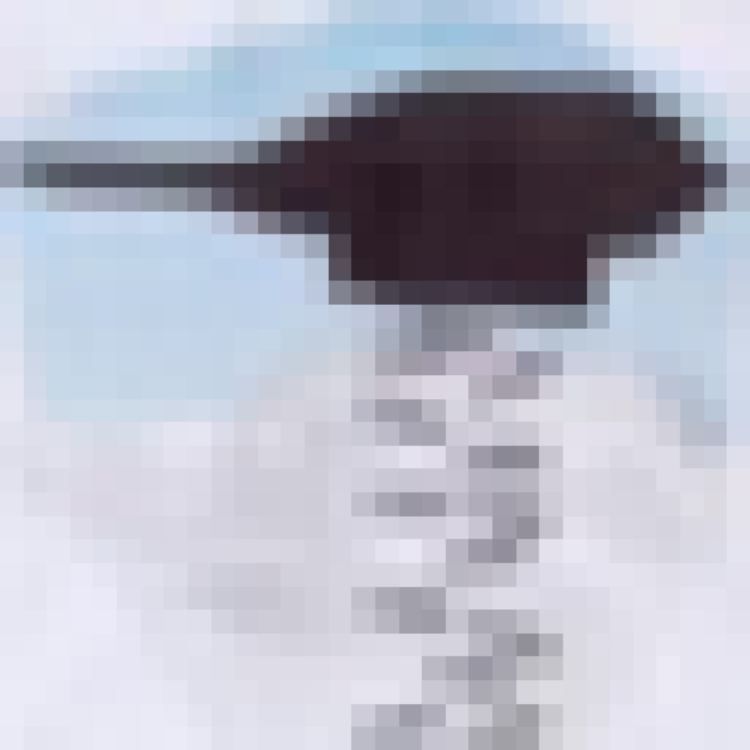}}
    \qquad
    \subfloat[$\mathcal{I}(x_i\Rightarrow x_t)=0.53012$]{\includegraphics[width=0.23\textwidth]{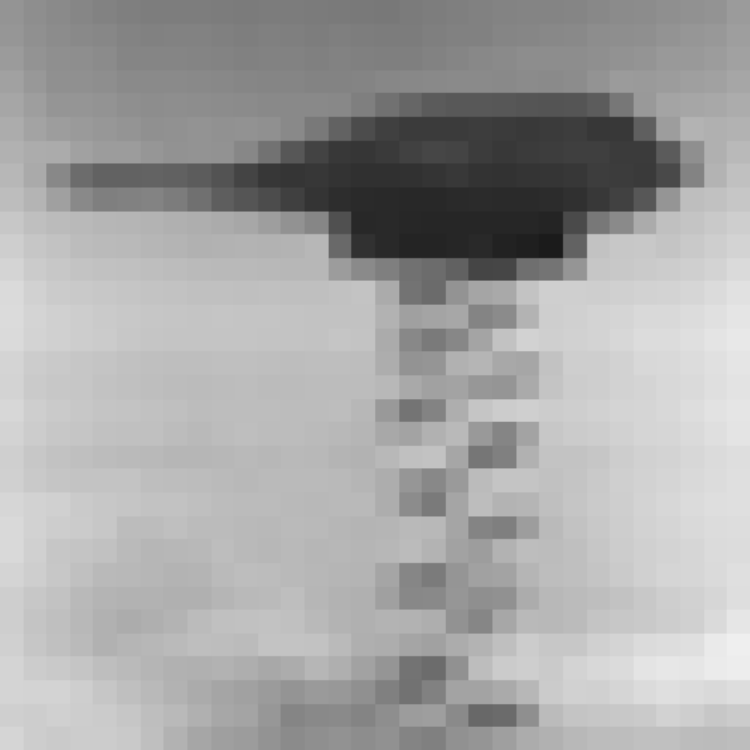}}
    \qquad
    \subfloat[Influence distribution]{\includegraphics[width=0.45\textwidth]{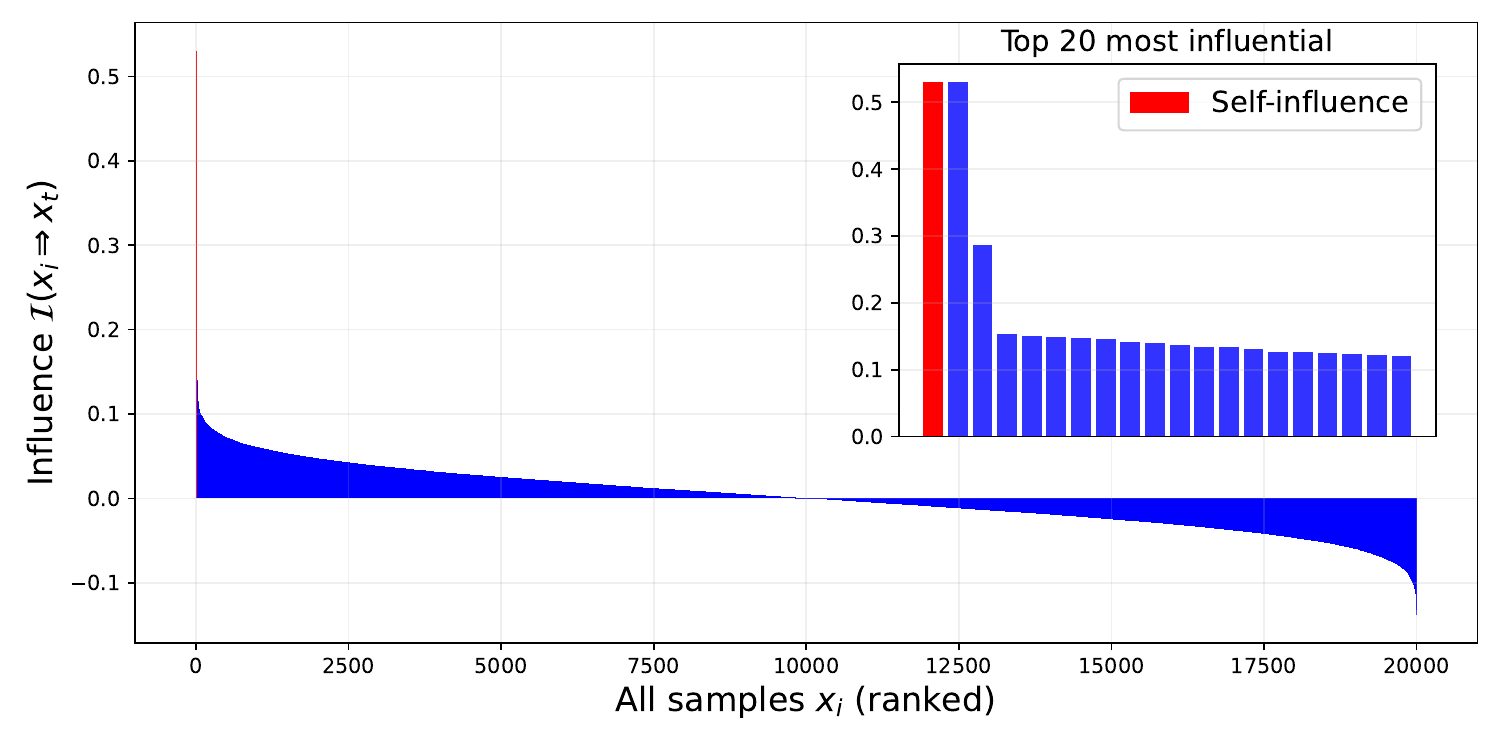}}
    \caption{Identifying near-duplicates in CIFAR-10 through the distribution of counterfactual influence. For each row, from left to right: (i) the target sample $x_t$ with its self-influence value $\mathcal{I}(x_t \Rightarrow x_t)$; (ii) the most influential sample different from the sample itself ($x_i \neq x_t$) with its influence value $\mathcal{I}(x_i \Rightarrow x_t)$; the full influence distribution for all $x_i$ for target record $x_t$. For all samples for which $\mathcal{I}(x_t \Rightarrow x_t)$ was larger that the median of all self-influence values, showing the $x_t$ with the \textbf{top 1-3} smallest Top-1 Influence Margin $\mathrm{IM}(x_t)$.}
    \label{fig:cifar_samples_1}
\end{figure*}

\begin{figure*}[ht]
    \centering
    \subfloat[$\mathcal{I}(x_t \Rightarrow x_t)=1.00965$]{\includegraphics[width=0.23\textwidth]{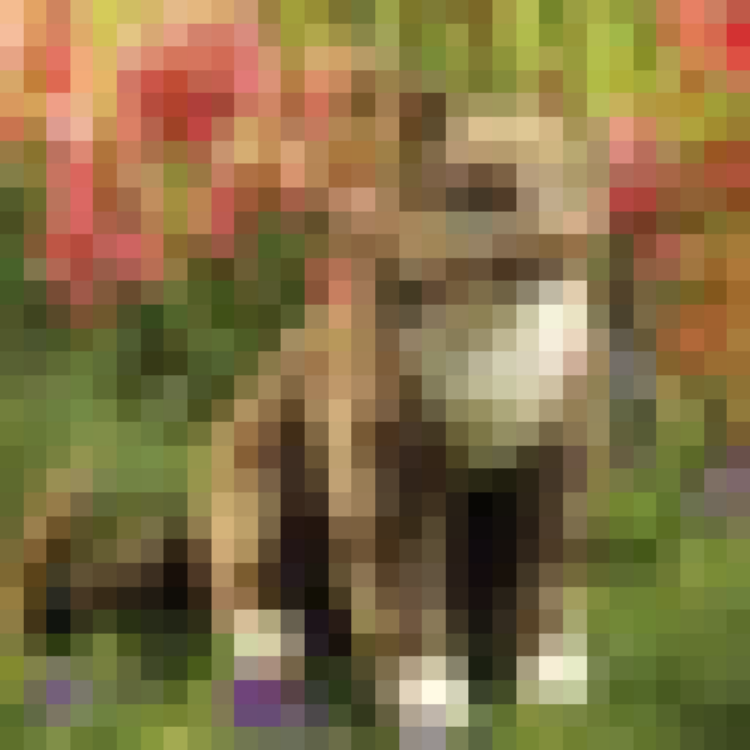}}
    \qquad
    \subfloat[$\mathcal{I}(x_i\Rightarrow x_t)=1.00702$ ]{\includegraphics[width=0.23\textwidth]{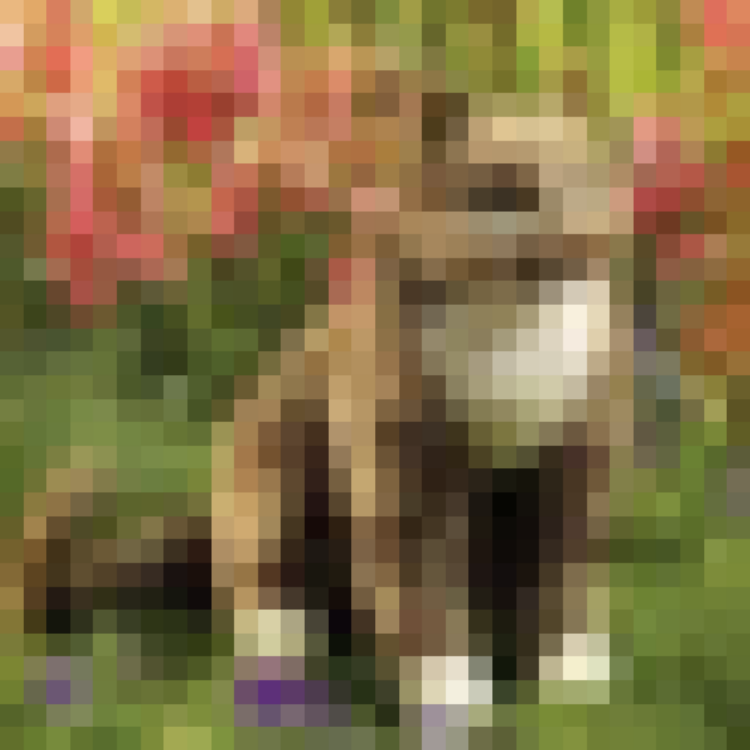}}
    \qquad
    \subfloat[Influence distribution]{\includegraphics[width=0.45\textwidth]{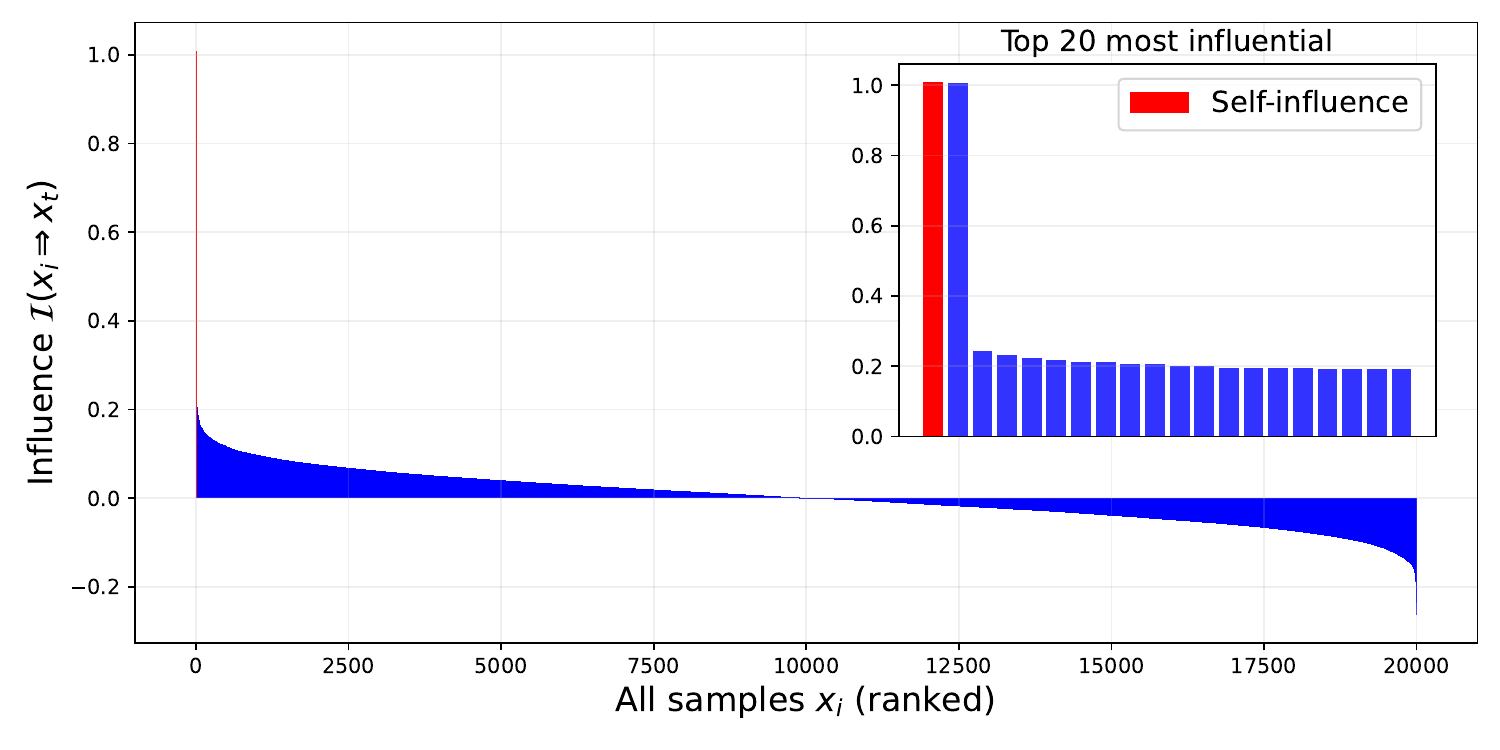}}
    \qquad
    \subfloat[$\mathcal{I}(x_t \Rightarrow x_t)=0.25191$]{\includegraphics[width=0.23\textwidth]{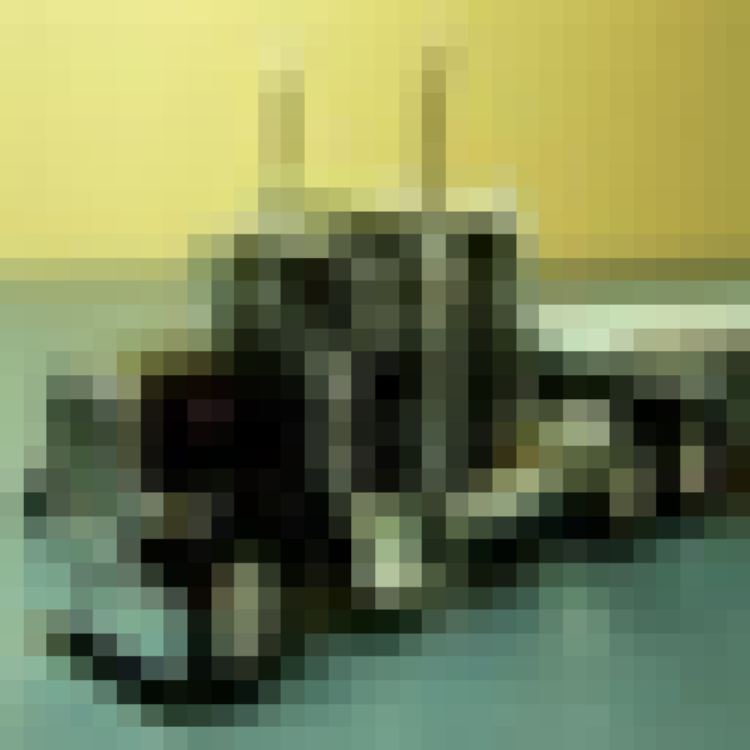}}
    \qquad
    \subfloat[$\mathcal{I}(x_i\Rightarrow x_t)=0.251156$ ]{\includegraphics[width=0.23\textwidth]{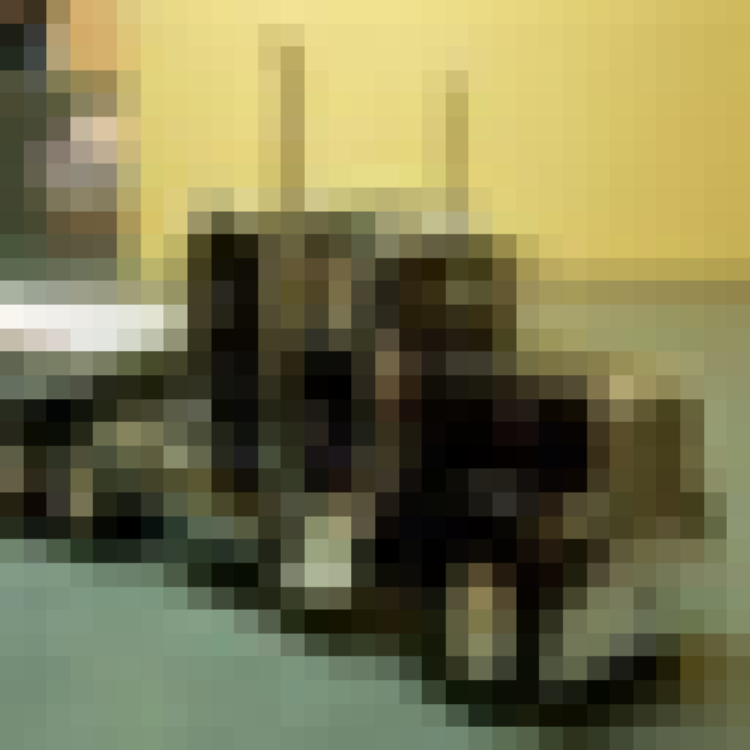}}
    \qquad
    \subfloat[Influence distribution]{\includegraphics[width=0.45\textwidth]{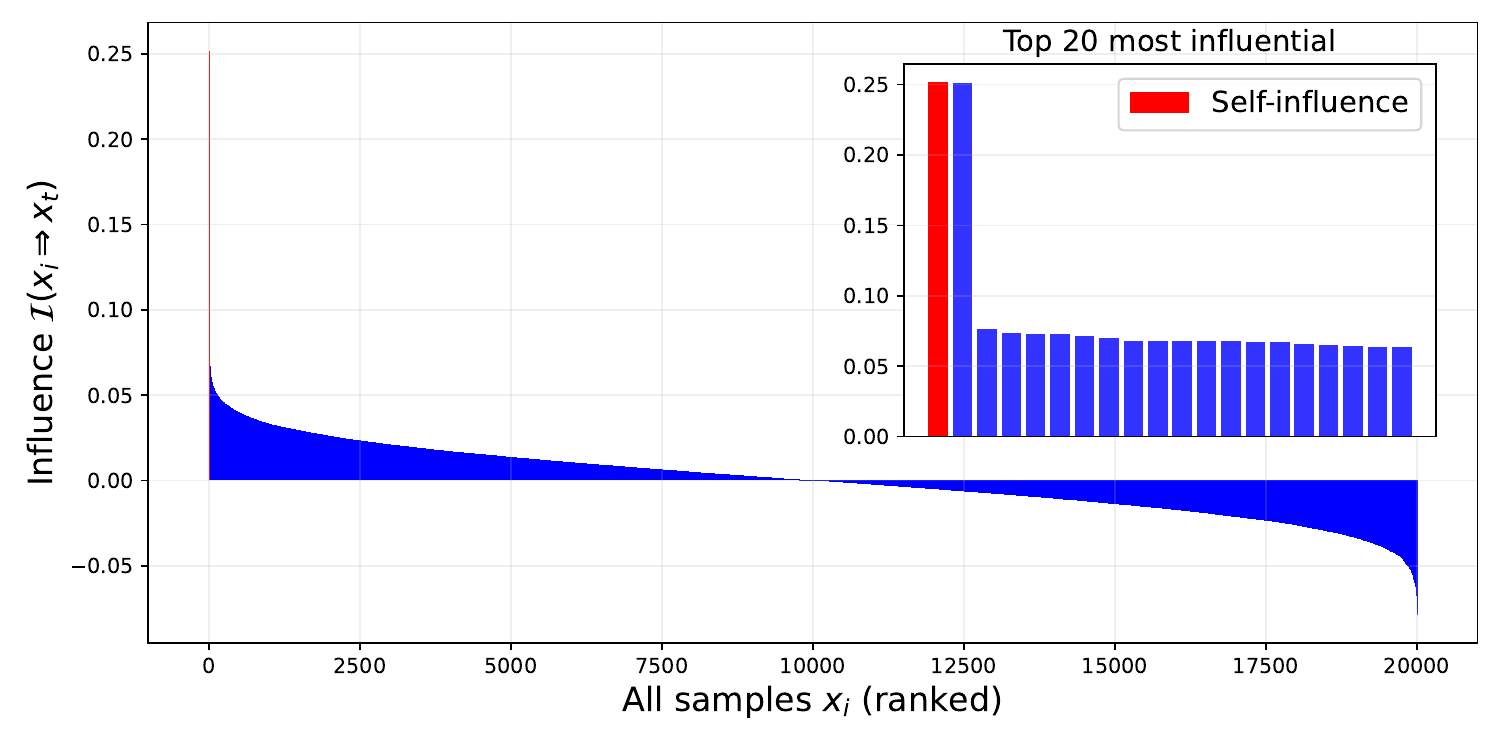}}
    \qquad
    \subfloat[$\mathcal{I}(x_t \Rightarrow x_t)=0.50944$]{\includegraphics[width=0.23\textwidth]{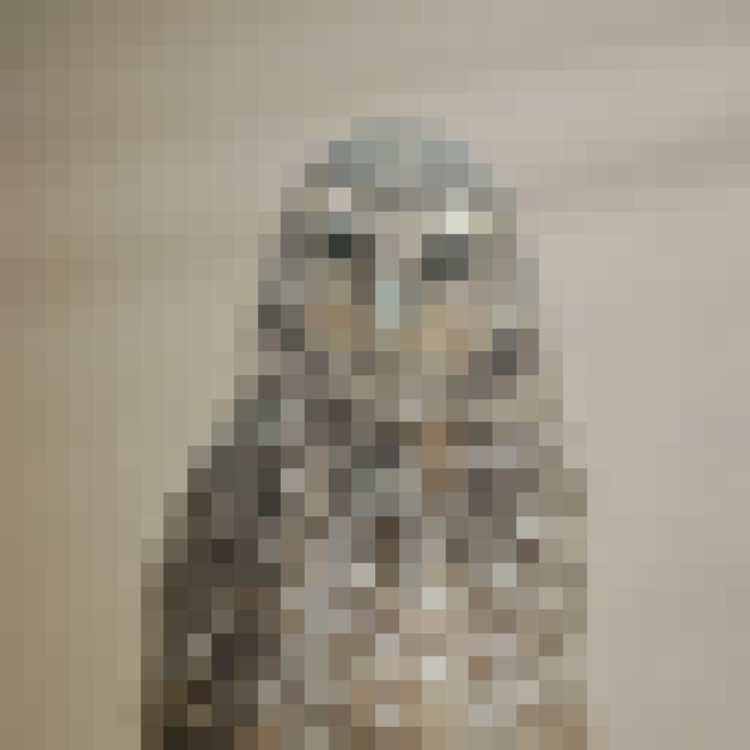}}
    \qquad
    \subfloat[$\mathcal{I}(x_i\Rightarrow x_t)=0.51317$ ]{\includegraphics[width=0.23\textwidth]{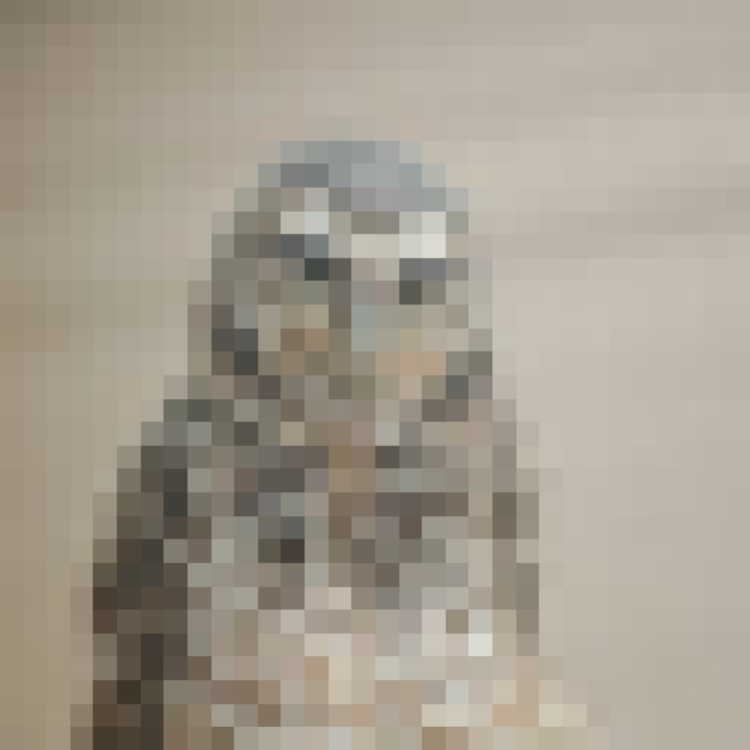}}
    \qquad
    \subfloat[Influence distribution]{\includegraphics[width=0.45\textwidth]{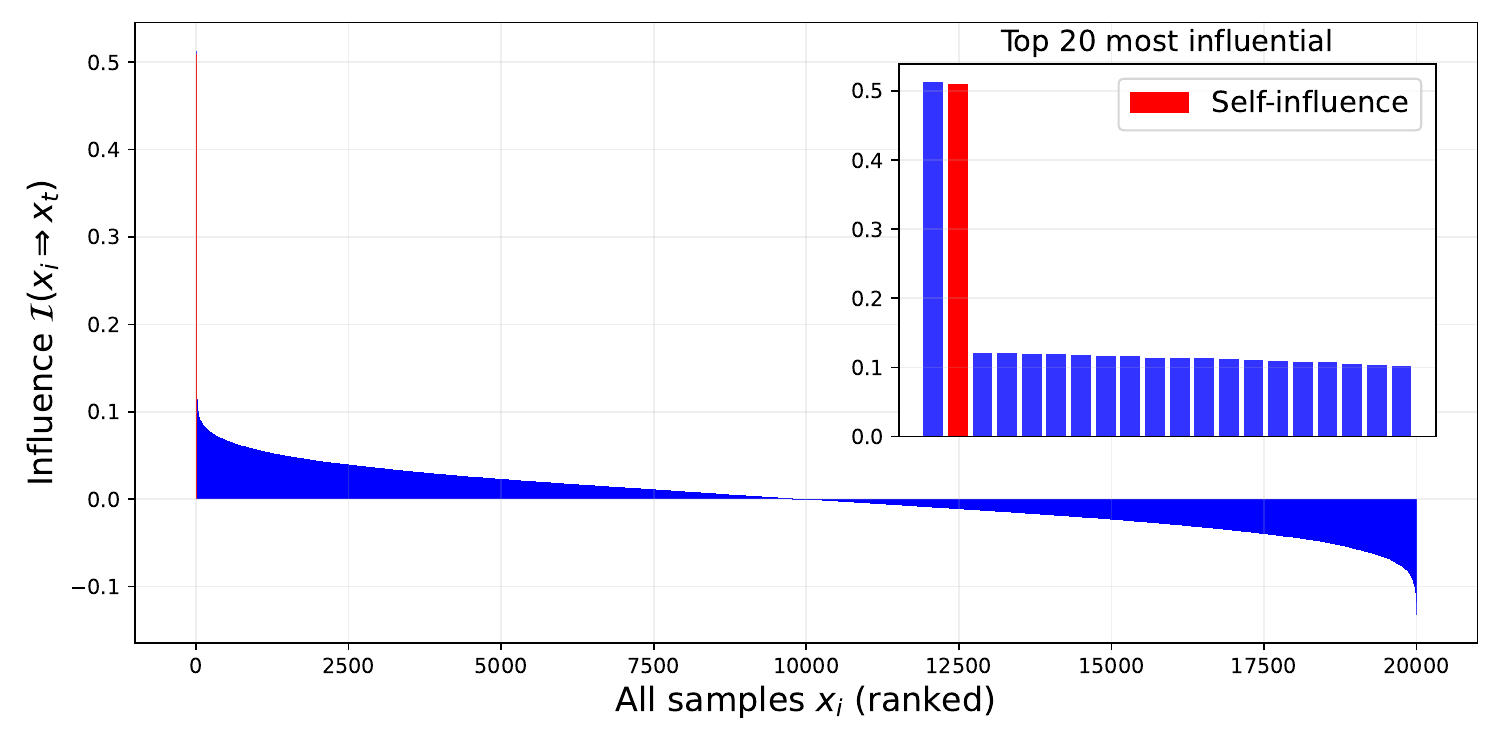}}
    \qquad
    \caption{Identifying near-duplicates in CIFAR-10 through the distribution of counterfactual influence. For each row, from left to right: (i) the target sample $x_t$ with its self-influence value $\mathcal{I}(x_t \Rightarrow x_t)$; (ii) the most influential sample different from the sample itself ($x_i \neq x_t$) with its influence value $\mathcal{I}(x_i \Rightarrow x_t)$; the full influence distribution for all $x_i$ for target record $x_t$. For all samples for which $\mathcal{I}(x_t \Rightarrow x_t)$ was larger that the median of all self-influence values, showing the $x_t$ with the \textbf{top 3-6} smallest Top-1 Influence Margin $\mathrm{IM}(x_t)$.}
    \label{fig:cifar_samples_2}
\end{figure*}

Figures~\ref{fig:cifar_samples_1} and~\ref{fig:cifar_samples_2} visualize samples with the smallest Top-1 Influence Margin $\mathrm{IM}(x_t)$, highlighting that for these samples, the most influential yet distinct sample appears to be a visual near-duplicate.  

\label{app:add_cifar_10}


\end{document}